\documentclass[10pt,journal,compsoc]{IEEEtran}

\ifCLASSOPTIONcompsoc
  \usepackage[nocompress]{cite}
\else
  \usepackage{cite}
\fi

\usepackage{xcolor}
\usepackage{graphicx}
\usepackage{amsmath}
\usepackage{acronym}
\usepackage{amssymb}
\usepackage{algorithmic}

\usepackage{subfigure}
\usepackage{subfloat}
\usepackage{multirow}
\usepackage{booktabs}

\usepackage{makecell}
\usepackage{array}
\usepackage{url}
\usepackage{commath}
\usepackage{textcomp}

\usepackage[pagebackref=true,breaklinks=true,colorlinks,bookmarks=false]{hyperref}
\usepackage{marvosym}

\usepackage[symbol]{footmisc}

\usepackage{xspace}
\usepackage{ragged2e}
\usepackage[labelfont=bf, font=footnotesize, justification=justified, format=plain]{caption}

\makeatletter
\DeclareRobustCommand\onedot{\futurelet\@let@token\@onedot}
\def\@onedot{\ifx\@let@token.\else.\null\fi\xspace}

\def\eg{\emph{e.g}\onedot} 
\def\ie{\emph{i.e}\onedot}

\def\etal{\emph{et al}\onedot}
\makeatother

\definecolor{ao}{rgb}{0.0, 0.5, 0.0}
\definecolor{blue}{rgb}{0.0, 0.0, 0.0} 
\newcommand{\name}{GTA-Human\xspace}
\newcommand{\cm}{\checkmark\xspace}

\begin{document}
\title{Playing for 3D Human Recovery}

\author{Zhongang Cai$^{\star}$ \quad
Mingyuan Zhang$^{\star}$ \quad
Jiawei Ren$^{\star}$ \quad
Chen Wei \quad
Daxuan Ren \\
Zhengyu Lin \quad
Haiyu Zhao \quad
Lei Yang \quad
Chen Change Loy \quad
Ziwei Liu$^{\href{mailto:ziwei.liu@ntu.edu.sg}{\textrm{\Letter}}}$%
\IEEEcompsocitemizethanks{
    \IEEEcompsocthanksitem $\star$ indicates equal contributions.
    \IEEEcompsocthanksitem Zhongang Cai, Mingyuan Zhang, Jiawei Ren, Daxuan Ren, Chen Change Loy and Ziwei Liu are with the S-Lab, Nanyang Technological University, Singapore, 639798.
    \IEEEcompsocthanksitem Zhongang Cai, Haiyu Zhao, Chen Wei, Zhengyu Lin, and Lei Yang are with Shanghai AI Laboratory.
    \IEEEcompsocthanksitem The corresponding author is Ziwei Liu: ziwei.liu@ntu.edu.sg
}%
\thanks{Manuscript received August 16, 2022.}}

\markboth{IEEE TRANSACTIONS ON PATTERN ANALYSIS AND MACHINE INTELLIGENCE}%
{Shell \MakeLowercase{\textit{et al.}}: Bare Advanced Demo of IEEEtran.cls for IEEE Computer Society Journals}


\IEEEtitleabstractindextext{%
\begin{abstract}
\justifying
Image- and video-based 3D human recovery (\ie, pose and shape estimation) have achieved substantial progress. However, due to the prohibitive cost of motion capture, existing datasets are often limited in scale and diversity.
In this work, we obtain massive human sequences by playing the video game with automatically annotated 3D ground truths. Specifically, we contribute \textbf{\name}, a large-scale 3D human dataset generated with the GTA-V game engine, featuring a highly diverse set of subjects, actions, and scenarios.
More importantly, we study the use of game-playing data and obtain five major insights.
\textbf{First}, game-playing data is surprisingly effective. A simple frame-based baseline trained on \name outperforms more sophisticated methods by a large margin. For video-based methods, \name is even on par with the in-domain training set.
\textbf{Second}, we discover that synthetic data provides critical complements to the real data that is typically collected indoor. \textcolor{blue}{We highlight that} our investigation into domain gap provides explanations for our data mixture strategies that are simple yet useful, \textcolor{blue}{which offers new insights to the research community.}
\textbf{Third}, the scale of the dataset matters. The performance boost is closely related to the additional data available. 
A systematic study on multiple key factors (such as camera angle and body pose) reveals that the model performance is sensitive to data density.
\textbf{Fourth}, the effectiveness of \name is also attributed to the rich collection of strong supervision labels (SMPL parameters), which are otherwise expensive to acquire in real datasets.
\textbf{Fifth}, the benefits of synthetic data extend to larger models such as deeper convolutional neural networks (CNNs) and Transformers, for which a significant impact is also observed.
We hope our work could pave the way for scaling up 3D human recovery to the real world. 
Homepage: \url{https://caizhongang.github.io/projects/GTA-Human/}.
\end{abstract}

}

\maketitle
\IEEEdisplaynontitleabstractindextext
\IEEEpeerreviewmaketitle

\ifCLASSOPTIONcompsoc
\IEEEraisesectionheading{\section{Introduction}\label{sec:introduction}}
\else
\section{Introduction}
\label{sec:introduction}
\fi


Image- and video-based 3D human recovery, \ie, simultaneous estimation of human pose and shape via parametric models such as SMPL \cite{loper2015smpl}, have transformed the landscape of holistic human understanding. This technology is critical for entertainment, gaming, augmented and virtual reality industries. However, despite that the exciting surge of deep learning is arguably driven by enormous labeled data \cite{krause2016unreasonable, sun2017revisiting}, the same is difficult to achieve in this field.
The insufficiency of data (especially in the wild) is attributed to the prohibitive cost of 3D ground truth (such as parametric model annotation) \cite{rong2019delving}. Existing datasets are either small in scale \cite{Sigal2009HumanEvaSV, von2018recovering, Leroy2020SMPLyB3}, collected in constrained indoor environment \cite{ionescu2013human3, Joo2019PanopticSA, Yu2020HUMBIAL}, or not providing the 3D parametric model annotation at all \cite{johnson2010clustered, Johnson2011LearningEH, lin2014microsoft, andriluka20142d}. 

\begin{figure}[t!]
  \centering
  \includegraphics[width=\linewidth]{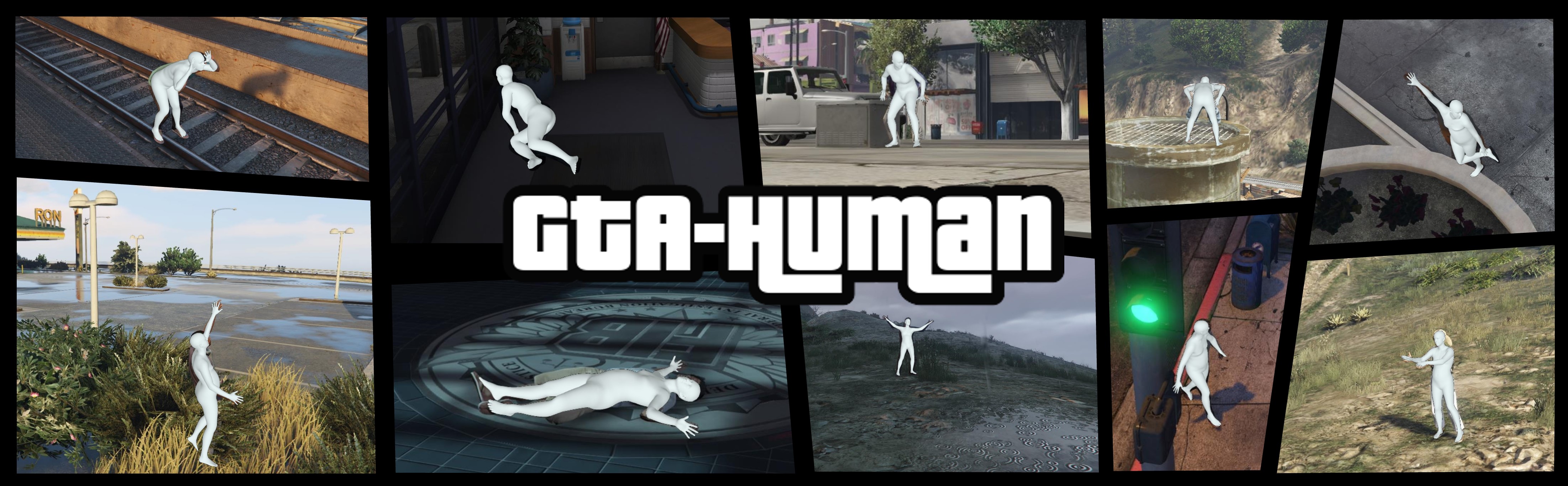}
  \caption{\textbf{\name} dataset is built from GTA-V, an open-world action game that features a reasonably realistic functioning metropolis and virtual characters living in it. Our customized toolchain enables large-scale collection and annotation of highly diverse human data that we hope aid in-depth studies on 3D human recovery. We show here a few examples with SMPL annotations overlaid on the virtual humans.}
\label{fig:teaser}
\end{figure}

Inspired by the success of training deep learning models with video game-generated data for various computer vision tasks such as instance segmentation \cite{richter2017playing}, 2D keypoint estimation \cite{fabbri2018learning}, motion prediction \cite{cao2020long}, mesh reconstruction \cite{hu2021sail}, detection and tracking \cite{fabbri2021motsynth}, we present \textbf{\name} (Figure \ref{fig:teaser}) in the hope to address the aforementioned limitations of existing datasets. \name is built by coordinating a group of computational workers (Figure \ref{fig:toolchain}) that simultaneously play the popular video game Grand Theft Auto V (GTA-V), to put together a large-scale dataset (Table \ref{tab:dataset_comparison}) with 1.4 million SMPL parametric labels automatically annotated in 20 thousand video sequences. Besides the scale, \name explores the rich resources of the in-game database to diversify the data distribution that is challenging to achieve in real life (Figure \ref{fig:data_diversity}, \ref{fig:actions} and \ref{fig:cam_angles}): more than \textit{600 subjects} of different gender, age, \textcolor{blue}{skin tone}, body shape and clothing; \textit{20,000 action clips} comprising a wide variety of daily human activities; \textit{six major categories of locations} with drastically different backgrounds from city streets to the wild; \textit{camera angles} are manipulated in each sequence to reflect a realistic distribution; subject-environment interaction that gives rises to \textit{occlusion} of various extents; time of the day that affects \textit{lighting} conditions, and \textit{weather} system that mimics the real climate changes. 

\begin{table*}[t]
  \begin{center}
  \caption{\textbf{3D human dataset comparisons.} We compare \name with existing real datasets with SMPL annotations and synthetic datasets with highly realistic setups. \name has competitive scale and diversity. Datasets are divided into three types: real, synthetic and mixed. \name samples character action sequences from a large in-game database that allows a unique action to be assigned to each video sequence. Note that EFT \cite{Joo2020ExemplarFF} re-annotates 2D human pose estimation datasets where the number of subjects are difficult to trace. *: 3DPW and Panoptic Studio only have general descriptions of scene activities.}
  \label{tab:dataset_comparison}
  \begin{tabular}{llcccrrrr}
    \hline
    Dataset & Year & Type & In-the-Wild & Video & 
    \#SMPL & \#Sequence & \#Subject & \#Action \\
    
    \hline
    
    
    
    

    HumanEva \cite{Sigal2009HumanEvaSV} & 2009 & Real & - & \cm &
    NA & 7 & 4 & 6 \\
    
    Human3.6M \cite{ionescu2013human3} & 2013 & Real & - & \cm &
    312K & 839 & 11 & 15 \\
    
    MPI-INF-3DHP \cite{mehta2017monocular} & 2017 & Mixed & \cm & \cm & 
    96K & 16 & 8 & 8 \\
    
    3DPW \cite{von2018recovering} & 2018 & Real & \cm & \cm &
    32K & 60 & 18 & * \\
    
    Panoptic Studio \cite{Joo2019PanopticSA} & 2019 & Real & - & \cm &
    736K & 480 & $\sim$100 & * \\
    
    EFT \cite{Joo2020ExemplarFF} & 2020 & Real & \cm & - &
    129K & NA & Many & NA \\
    
    SMPLy \cite{Leroy2020SMPLyB3} & 2020 & Real & \cm & \cm &
    24K & 567 & 742 & NA \\ 
    
    AGORA \cite{Patel:CVPR:2021} & 2021 & Synthetic & \cm & - & 
    173K & NA & $>$350 & NA  \\
    
    \textbf{\name} & 2022 & Synthetic & \cm & \cm &
    1.4M & 20K & $>$600 & 20K \\
    
    \hline
  \end{tabular}
\end{center}
\end{table*}

Equipped with \name, we conduct an extensive investigation in the use of synthetic data for 3D human recovery. 
\textbf{1)} \textbf{Better 3D human recovery with data mixture.} Despite the seemingly unavoidable domain gaps, we show that practical settings that mix synthetic data with real data, such as blended training and pretraining followed by finetuning, are surprisingly effective. 
First, HMR \cite{kanazawa2018end}, one of the first deep learning-based methods for SMPL estimation with relatively simplistic architecture, when trained with data mixture, is able to outperform more recent methods with sophisticated designs or additional information such as SPIN \cite{kolotouros2019learning} and VIBE \cite{kocabas2020vibe}.
Moreover, PARE \cite{kocabas2021pare}, a state-of-the-art method also benefit considerably from \name. 
Second, our experiments on the video-based method VIBE \cite{kocabas2020vibe} further demonstrate the effectiveness of data mixture: an equal amount of synthetic \name data is as good as a real-captured indoor dataset as the frame feature extractor is already pretrained on real datasets; the full set of \name is even on par with in-domain training data. 

\textbf{2)} \textbf{Closing the domain gap with synthetic data.}
\textcolor{blue}{We conduct a pioneering study on the root causes behind the effectiveness of game-playing data}. An investigation into the domain gaps provides insights into the complementary nature of synthetic and real data: despite the reality gap, the synthetic data embodies the diversity that many real datasets (typically collected indoor) lack. Moreover, we experiment with mainstream domain adaptation methods to further close the domain gaps and obtain improvements.

\textbf{3)} \textbf{Dataset scale matters.} We demonstrate that adding game-playing data progressively improves the model performance. Considering the difficulty of collecting real data with ground truth 3D annotations, synthetic data may thus be an attractive alternative. Moreover, a multi-factor analysis reveals that supervised learning leads to severe sensitivity to data density. Amongst factors such as camera angles, pose distributions, and occlusions, a consistent drop in performance is observed where data is scarce. \textcolor{blue}{Hence, our observation suggests that synthetic datasets can play a vital role in supplementing corner case scenarios in real practice}.

\textbf{4)} \textbf{Strong supervision (SMPL) is key.} Compared to large-scale pose estimation benchmarks that only provide 3D keypoints, we demonstrate that strong supervision in the form of SMPL parameters may be quintessential for training a strong model. \textcolor{blue}{In a greater depth than prior arts, we discuss the potential reasons and reaffirm the value of \name as a scalable training source with SMPL annotations.} 

\textbf{5)} \textbf{Big data benefits big models.} Despite recent development in deeper convolutional networks \cite{he2016deep, xie2017aggregated} and vision transformers \cite{dosovitskiy2020image, touvron2020training} in computer vision research, the mainstream backbone size remains unchanged for 3D human recovery \cite{kanazawa2018end, kolotouros2019learning, kocabas2020vibe}. We extend our study to deeper CNNs and Transformers, and show that training with \name 
enables performance boosts for both small and large backbones. Interestingly, smaller, backbones trained with additional GTA-Human could outperform larger counterparts trained with only real data.
\section{Related Work}

\subsection{3D Human Recovery}

\noindent \textbf{Human Parametric Models.}
Unlike human pose estimation that uses skeleton (joint keypoints) to represent humans~\cite{martinez2017simple, pavllo20193d}, human pose and shape estimation is typically performed with 3D human parametric models, such as SMPL \cite{loper2015smpl}, SMPL-X \cite{pavlakos2019expressive} and STAR \cite{osman2020star}, which take in parameters that represent pose and shape of the human subject, and output 3D human mesh via linear blend skinning. We base our discussion on SMPL version 1.0 in this work that consists of pose parameters $\theta \in \mathbb{R}^{72}$ and shape parameters $\beta \in \mathbb{R}^{10}$, for its popularity.

\vspace{2mm}
\noindent \textbf{Registration-based Methods.}
As the output of the human parametric model is manipulated by body parameters, SMPLify \cite{bogo2016keep} and the following SMPLify-X \cite{pavlakos2019expressive} are the pioneering works to optimize these parameters to minimize the distance between ground truth 2D keypoints and reprojected human mesh joints. SMPLify is also extended to videos with temporal constraints employed \cite{huang2017towards}. Although optimization-based methods are able to achieve impressive results, they are slow and typically take more than 60 seconds per frame. Hence, recent work \cite{fan2021revitalizing} has been proposed to accelerate optimization.

\vspace{2mm}
\noindent \textbf{Regression-based Methods.}
Direct regression of body parameters using a trained deep learning model has gained more popularity due to fast inference. The recent works are categorized into image-based \cite{pavlakos2018learning, omran2018neural, guler2019holopose, georgakis2020hierarchical, li2020hybrik, sun2020monocular, kocabas2021pare}, and video-based \cite{kanazawa2019learning, sun2019human, mehta2020xnect, moon2020i2l, choi2021beyond, luo20203d} methods. HMR \cite{kanazawa2018end} is a pioneering end-to-end deep learning-based work, which takes ResNet-50 \cite{he2016deep} as its backbone and directly regress the parameters of $\theta$ and $\beta$. VIBE \cite{kocabas2020vibe} is a milestone video-based work that leverages temporal information for realistic pose sequences. Recently, a transformer encoder is introduced for vertex-joint reweighting \cite{lin2020end}, but the method still uses a CNN backbone for feature extraction.

\vspace{2mm}
\noindent \textbf{Mixed Methods.}
There is a line of work that combines optimization-based and regression-based techniques. SPIN \cite{kolotouros2019learning}, adds a SMPLify step to produce pseudo parametric labels to guide the learning of the network. SPIN address the lack of SMPL annotation but the optimization step results in slow training. Others propose to refine the per-frame regression results by bundle adjustment of the video sequence as a whole \cite{arnab2019exploiting}, designing a new swing-twist representation to replace the original axis-angle representation of SMPL \cite{li2020hybrik}, and finetuning a trained network to obtain refined prediction \cite{Joo2020ExemplarFF}, and employing a network to predict a parameter update rule in iterations of optimization \cite{song2020human}.

\subsection{Datasets}

\noindent\textbf{Datasets with 2D Keypoint Annotations.}
Many datasets contain in-the-wild images, albeit the lack of SMPL annotations, they provide 2D keypoint labels. Datasets such as LSP \cite{johnson2010clustered}, LSP-Extended \cite{Johnson2011LearningEH}, COCO \cite{lin2014microsoft} and MPII \cite{andriluka20142d} contain images crawled from the Internet, and are annotated with 2D keypoints manually. Such a strategy allows a large number of in-the-wild images to be included in the dataset. To obtain 3D annotations that are crucial to human pose and shape estimation, a common method is to fit an SMPL model on 2D keypoints. SSP-3D \cite{Sengupta2020SyntheticTF} and 3DOH50K \cite{Zhang2020ObjectOccludedHS} leverages pre-trained model to perform keypoint estimation as the first step, whereas UP-3D \cite{lassner2017unite} and EFT \cite{Joo2020ExemplarFF} performs fitting on ground truth keypoints. However, these datasets typically suffer from the inherent depth ambiguity of images and the pseudo-SMPL may not have the accurate scale.

\vspace{2mm}
\noindent\textbf{Real Datasets.}
Motion capture facilities are built to achieve high-accuracy 3D annotations. HumanEva \cite{Sigal2009HumanEvaSV} and Human3.6M \cite{ionescu2013human3} employ optical motion capture systems, but intrusive markers are needed to be placed on the subjects. Total Capture \cite{Trumble2017TotalC3} MuPoTS-3D \cite{Mehta2018SingleShotM3}, Panoptic Studio \cite{Joo2019PanopticSA}, and HUMBI \cite{Yu2020HUMBIAL} make use of multiple camera views and require no intrusive marker. However, the background is constant and thus lacks diversity. 3DPW \cite{von2018recovering} combines inertial measurement units (IMUs) and a moving camera to build an in-the-wild dataset with 3D annotations. 3DPW has become an important benchmark for 3D human recovery. Nevertheless, the IMU drift is still an obstacle and the dataset only contains a relatively small number of videos. 
SMPLy \cite{Leroy2020SMPLyB3} constructs point clouds from multi-view capture of static people and fits SMPL on them. However, the scale of the dataset is limited by the difficulty of collecting videos that meet the special setup requirement. HuMMan \cite{cai2022humman} is the most recent large-scale multi-modal 4D human dataset.

\vspace{2mm}
\noindent\textbf{Synthetic or Mixed Datasets.}
SURREAL \cite{Varol2017LearningFS}, Hoffmann \etal  \cite{Hoffmann2019LearningTT} render textured SMPL body models in real-image backgrounds. However, this strategy does not account for the geometry of the clothes, where the mismatch may result in unrealistic subjects. 3DPeople \cite{Pumarola20193DPeopleMT} uses clothed human models while MPI-INF-3DHP \cite{mehta2017monocular} takes segmented subjects from images and paste them onto new backgrounds in the training set. However, the subject-background interaction is still unnatural. AGORA \cite{Patel:CVPR:2021} is a recent synthetic dataset featuring high-quality annotations by rendering real human scans in a virtual world. However, the dataset is image-based and does not support the training of video-based methods.
Richter \etal \cite{richter2016playing, richter2017playing}, Kr{\"a}henb{\"u}hl \etal \cite{krahenbuhl2018free}, JTA \cite{fabbri2018learning}, GTA-IM \cite{cao2020long}, SAIL-VOS 3D \cite{hu2021sail}, MOTSynth \cite{fabbri2021motsynth} have demonstrated the potential of obtaining nearly free and perfectly accurate annotations from video games for various computer vision tasks. Amongst them, JTA provides 3D keypoint for pedestrian pose estimation (in the form of keypoints) and tracking, SAILVOS3D focuses on object detection and mesh reconstruction (including non-parametric human meshes). However, these datasets do not provide SMPL annotation needed for our investigation. We take inspiration from these works in building \name specifically for human pose and shape estimation via parametric regression.

\section{\name Dataset}
\label{sec:dataset}
\begin{figure*}[t!]
  \centering
  \includegraphics[width=\linewidth]{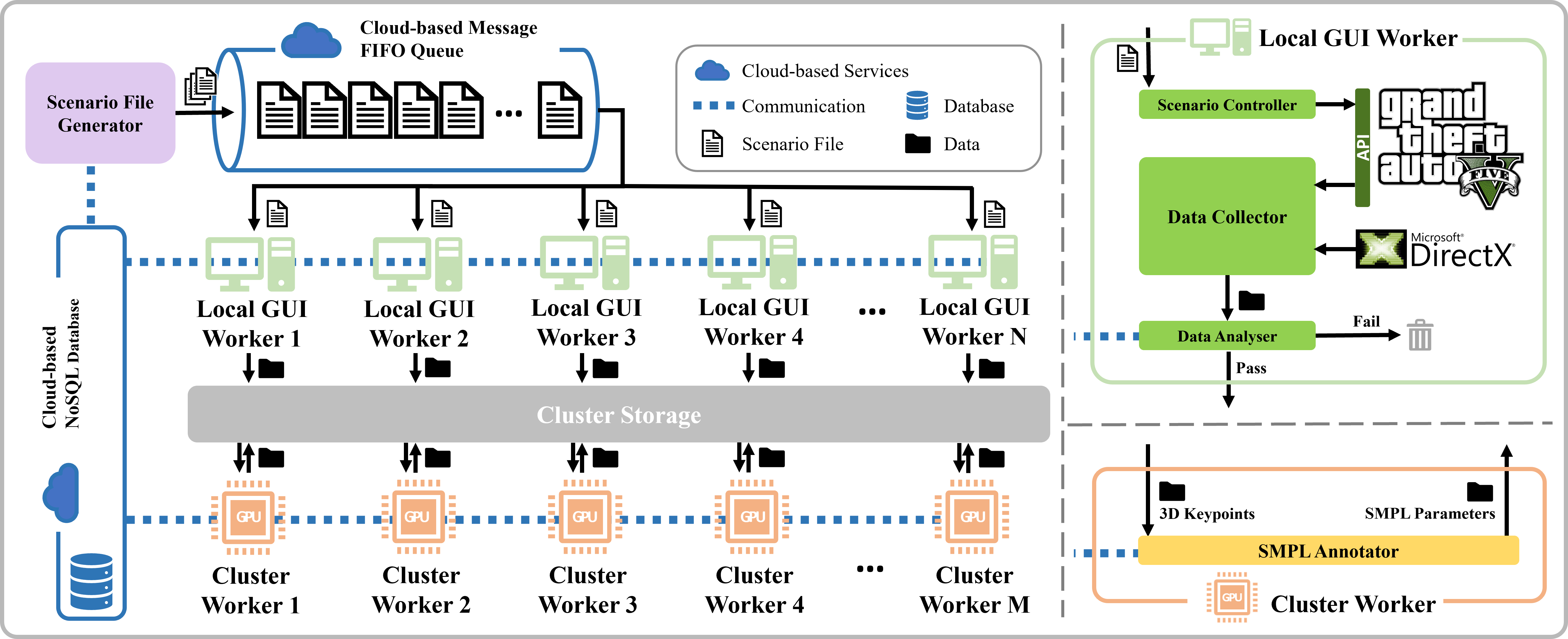}
  \caption{\textbf{Data collection toolchain.} Our toolchain is highly scalable as the cloud services are used to coordinate a large number of computation workers. Left: the overview of the pipeline. Top right: an elaborate illustration of Local GUI Worker. Bottom right: an elaborate illustration of Cluster Worker. }
  \label{fig:toolchain}
\end{figure*}

A scale comparison between \name and existing dataset is shown in Table \ref{tab:dataset_comparison}. \name features 1.4 million individual SMPL annotations, which is highly competitive compared to other real datasets and synthetic datasets with realist setups.
Moreover, GTA-Human consists of diverse (especially outdoor) scenes that are expensive and difficult to collect accurate annotation in real life.
Notably, \name provides video sequences instead of static frames and supports video-based human recovery.

\subsection{Toolchain}
\label{sec:dataset:toolchain}

\begin{figure*}[t!]
  \centering
  \includegraphics[width=\linewidth]{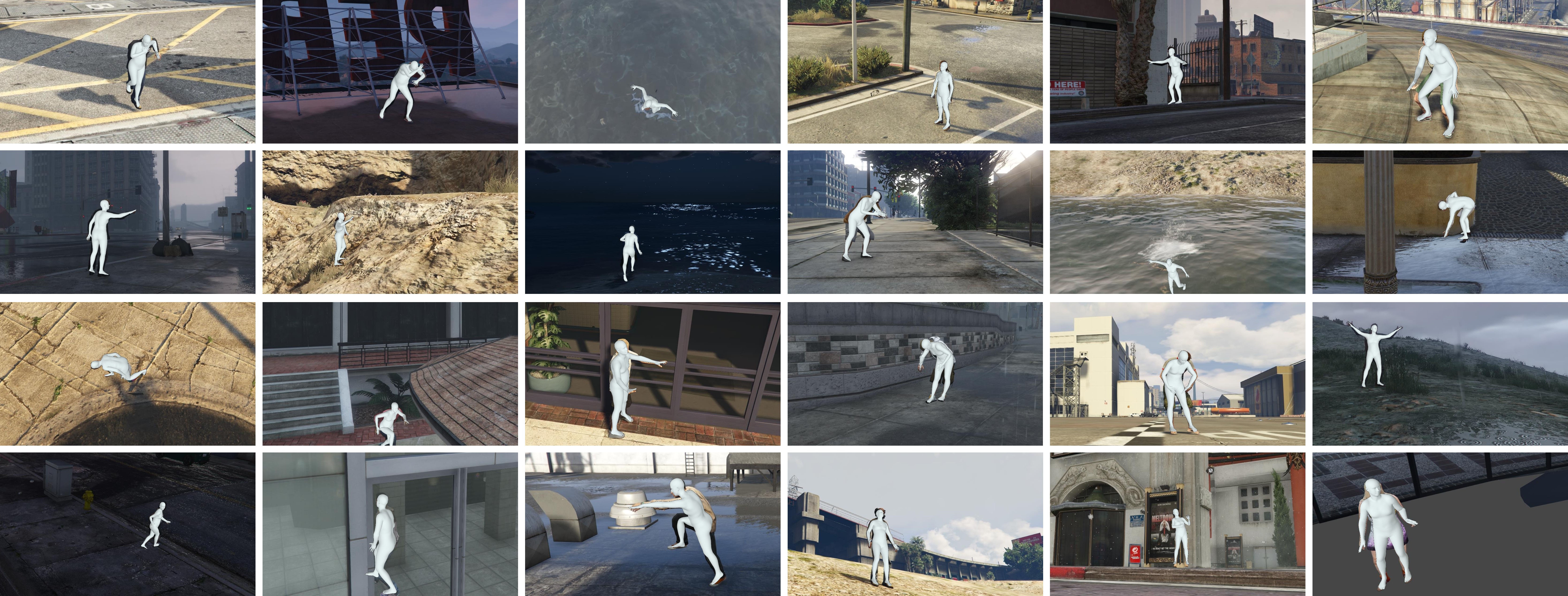}
  \caption{\textbf{More examples of \name.} We highlight that \name is a large-scale, highly diverse (in terms of factors such as subjects, actions, locations, and camera angles) dataset. Each frame of the video clips is annotated with SMPL parameters.}
  \label{fig:more_visualizations}
\end{figure*}

Inspired by existing works that use GTA-generated data for various vision tasks \cite{richter2016playing, richter2017playing, krahenbuhl2018free, fabbri2018learning, cao2020long, hu2021sail, fabbri2021motsynth}, our toolchain extracts ground truth 2D and 3D keypoints, semantic and depth maps from the game engine, followed by fitting SMPL models on the keypoints with temporal constraints. To achieve scalability and efficiency, we design and deploy an automatic system (Fig. \ref{fig:toolchain}) that leverages cloud-based services for parallel deployment and coordination of our tools on a large number of computer instances and GPU cluster nodes. \name consists of sequences of single-person scenes. More examples in \name are found in Fig. \ref{fig:more_visualizations}. 

\vspace{2mm} 
\noindent \textbf{Cloud-based NoSQL Database.}
The unit of data in \name is a single video sequence. Hence, we employ a Database that is hosted on the cloud, to track the progress of data generation and processing of each sequence. The status of a sequence is updated at each stage in the toolchain, which we elaborate on the details below.

\vspace{2mm} 
\noindent \textbf{Scenario File Generator.}
This tool reads from the Database to retrieve sequence IDs that are either not generated before or failed in previous processing attempts, and produce random scene attributes such as subject ID, action ID, location in the 3D virtual world, camera position and orientation, lighting, and weather settings. 

\vspace{2mm} 
\noindent \textbf{Cloud-based Message FIFO Queue.}  The Message FIFO Queue parse the scenario files from the Scenario File Generator as text strings, which can be fetched in the first-in-first-out (FIFO) manner by multiple Local GUI Workers. Note that the queue allows for multiple workers to retrieve their next jobs simultaneously.

\vspace{2mm} 
\noindent \textbf{Local GUI Workers.}
We purchase multiple copies of GTA-V and install them on regular gaming desktops. We refer to these desktops as Local GUI Workers. Each worker runs three tools: Scenario Controller, Data Collector, and Data Analyser which we elaborate on below.

\vspace{2mm} 
\noindent \textbf{Scenario Controller.}
Taking scenario files as the input, Scenario Controller is essentially a plugin that interacts with the game engine via the designated Application Programming Interface (API). It is thus able to control the subject generation and placement, action assignment to the subject, camera placement, in-game time, and weather.

\vspace{2mm} 
\noindent \textbf{Data Collector.}
This tool obtains data and some annotations from the API provided by GTA-V. First, it extracts 3D keypoints from each subject via the API provided by GTA-V. In addition to the original 98 3D keypoints available, we further obtain head top \cite{fabbri2018learning} and nose from interpolation of existing keypoints. 
We used the perspective camera model as we have access to the intrinsic and extrinsic parameters of the virtual camera and 3D keypoints of the subjects in world coordinates. Hence, we transform the 3D keypoints in the camera coordinates. We then project 3D keypoints to the image plane to obtain 2D keypoints.
Second, we project light rays at each joint to determine if the joint is occluded or self-occluded by checking the entity that the light ray hits first \cite{fabbri2018learning}. Third, our tool intercepts the rendering pipeline, powered by DirectX, for depth maps and semantic masks. The pixel-wise depth is directly read from depth buffers. Shader injection enables the segmentation of individual patches, and we manually assign the semantic class to various shaders based on their variable names. We refer interested readers to \cite{krahenbuhl2018free} for more details. Fourth, the collector also records videos. 
 
\vspace{2mm} 
\noindent \textbf{Data Analyser.}
To filter out low-quality data in the early stage, Data Analyser imposes several constraints on 3D keypoints obtained. We compute joint movement speed simply as the position different in consecutive frames to filter out less expressive actions (slow-moving or stationary actions). Severely occluded, or out-of-view subjects are also flagged at this stage. If sequences pass the analysis, their data are transferred from the local storage to a centralized storage space on our GPU cluster  (Cluster Storage) for further processing. The failed ones, however, are deleted. The Database is notified of the result to get the status updated.

\vspace{2mm} 
\noindent \textbf{Cluster Workers and SMPL Annotator.}
On each Cluster Worker (a GPU in the cluster), we run an instance of SMPL Annotator that takes keypoint annotation from the Cluster Storage. We upgrade SMPLify \cite{bogo2016keep} in two ways to obtain accurate SMPL annotation. 1) we find out that compared to 2D keypoints that have inherent depth ambiguity, exacerbated by weak perspective projection \cite{kissos2020beyond}, 3D keypoints are unambiguous. Minor modifications are needed to replace the 2D keypoint loss of the original SMPLify with 3D keypoint loss. 2) Taking advantage of the fact that \name consists of video sequences instead of unrelated images, temporal consistency in the form of rotation smoothing and unified shape parameters are enforced. The SMPL parameters include $\theta$ and $\beta$, and an additional translation vector, are optimized at an average of one second per frame. We visualize more examples in \name that are produced with our SMPL annotation tool in Fig. \ref{fig:more_visualizations}.

\subsection{Data Diversity}
\label{sec:dataset:data_diversity}
\begin{figure*}[t!]
  \centering
  \includegraphics[width=\linewidth]{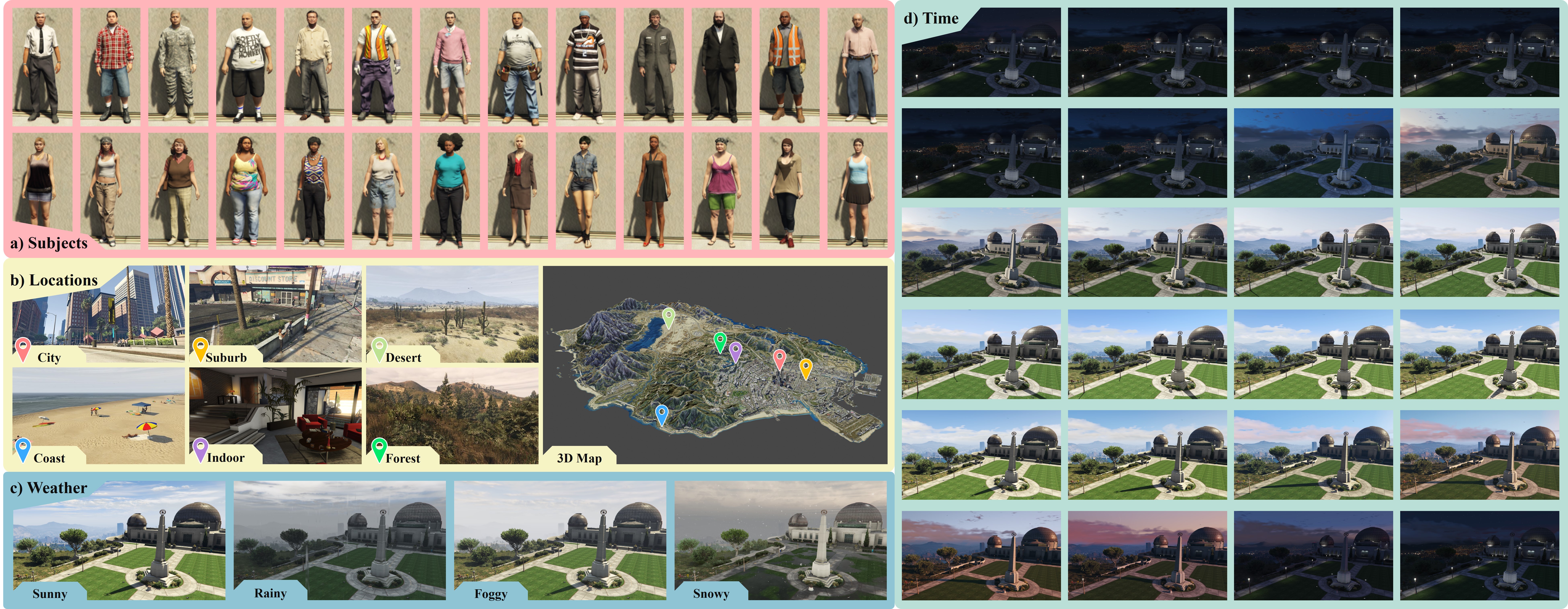}
  \caption{\textbf{Data diversity in \name.} 
  \textbf{(a)} \name contains subjects of varied genders, ages, \textcolor{blue}{skin tones}, clothing and body shapes. 
  \textbf{(b)} locations with diverse backgrounds.  
  The example locations are pinpointed on the 3D game world map. We discover in Section \ref{sec:experiments:the_unreasonable_effectiveness_of_data} that the outdoor scenes are critical to the usefulness of \name.
  \textbf{(c)} Different weather conditions.
  \textbf{(d)} In-game time is set to capture diverse lighting conditions. We capture the same scene at one game hour interval. Note the shadow direction is affected by the sun's position.}
  \label{fig:data_diversity}
\end{figure*}

\begin{figure}[t]
  \centering
  \includegraphics[width=\linewidth]{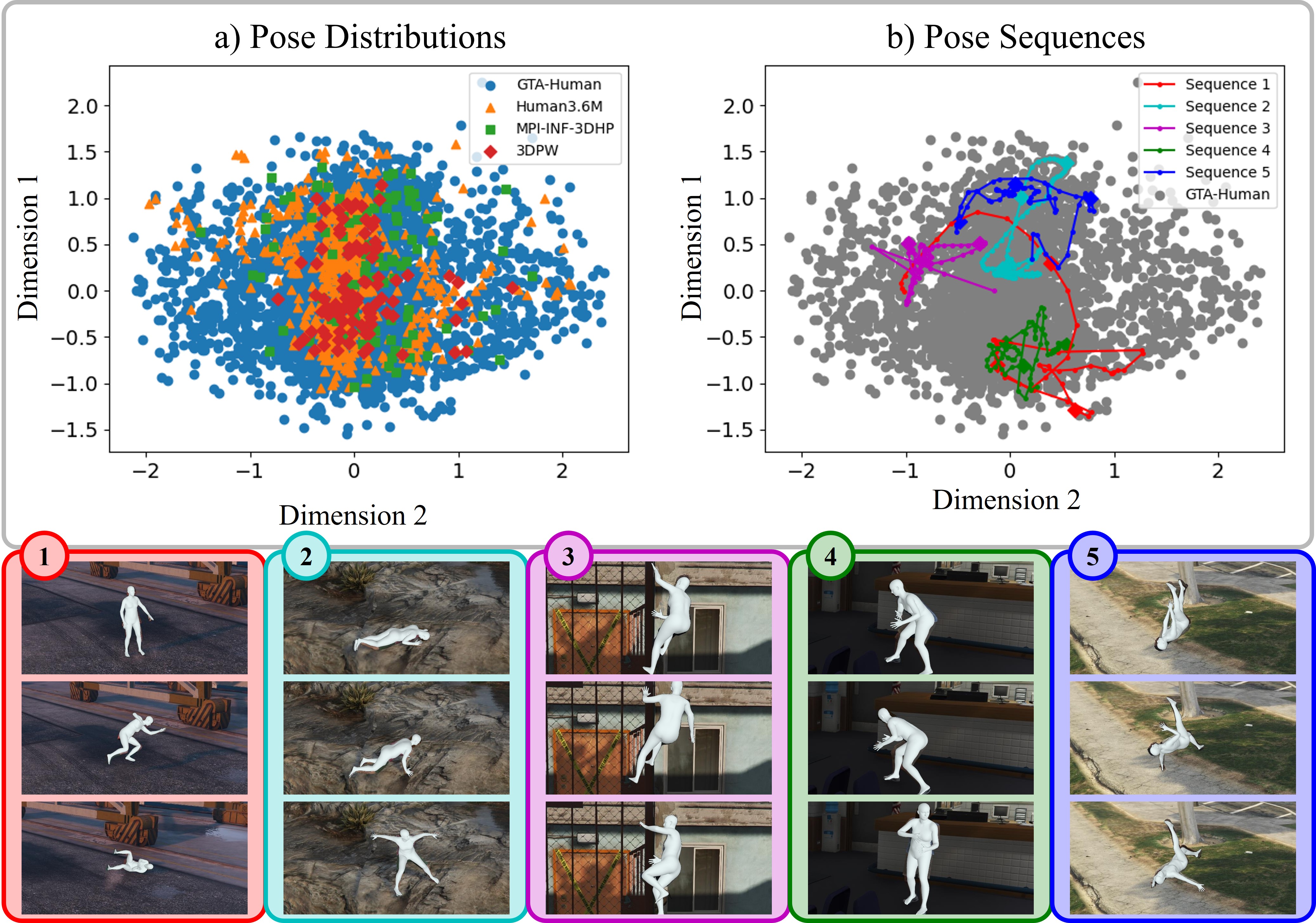}
  \caption{\textbf{Actions.} \name contains 20 thousand actions that are expressive and diverse. \textbf{(a)} The distribution of poses in \name and real datasets are visualized after PCA dimension reduction.  \textbf{(b)} We show five pose sequences, represented by curves. Representative frames of sequence 1-5 are indicated by the diamond-shaped nodes. Datasets are downsampled proportionally.}
  \label{fig:actions}
\end{figure}

\begin{figure}[t]
  \centering
  \vspace{-2mm}
  \includegraphics[width=\linewidth]{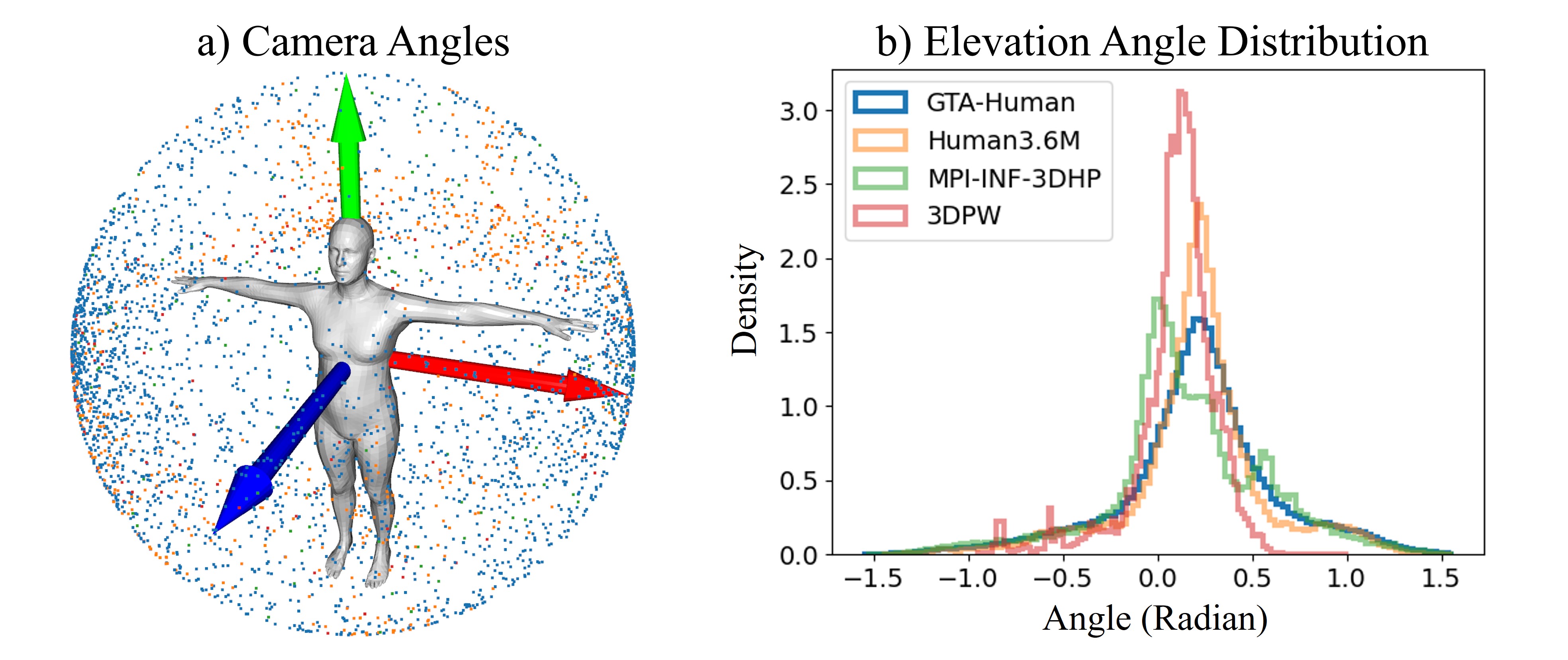}
  \caption{\textbf{Camera angles.} \textbf{(a)} Visualization of camera angles sampled from various datasets, normalized to a unit sphere. \textbf{(b)} Elevation angle (up-down, with positive value indicating a camera placed higher than the waist and looking down) distributions. The vertical axis represents normalized data density.
  The colors of the points in (a) and line plots in (b) represent different datasets, shown in the legend in (b).}
  \label{fig:cam_angles}
\end{figure}

Due to the difficulty and cost of data collection and SMPL annotation for the 3D human recovery task, most existing datasets are built at restrictive locations such as indoor studios or laboratory environments. Furthermore, only a small number of subjects are usually employed to perform a limited set of actions. In contrast, \name is designed to maximize the variety in the following aspects. We demonstrate the diversity in subjects, locations, weather, and time (light conditions) in Fig. \ref{fig:data_diversity}, actions in Fig. \ref{fig:actions}, and camera angles in Fig. \ref{fig:cam_angles}.

\vspace{2mm}
\noindent\textbf{Subjects.}
\name collects over 600 subjects of different genders, ages, \textcolor{blue}{skin tones}, clothing, and body shapes for a wide coverage of human appearances. In addition, unlike motion capture systems in real life that rely on intrusive markers to be placed on the subjects, accurate skeletal keypoints are obtained directly from the game's API.

\noindent\textbf{Actions.}
Existing datasets either design a small number of actions \cite{Sigal2009HumanEvaSV, ionescu2013human3, mehta2017monocular}, or lack a clearly defined action set \cite{von2018recovering, Joo2019PanopticSA}. In contrast, we gain access to a large database of motion clips (actions) that can be used to manipulate the virtual characters, whose typical length is 30-80 frames at 30 FPS. These actions provide a fairly holistic representation of city-dwellers' daily activities, and are reasonably realistic because they are originally produced via motion capture of real human actors or actresses. We select 20,000 most dynamic and expressive actions. In Fig. \ref{fig:actions}, the distribution of \name poses does not only have the widest spread, but also covers existing poses in the real datasets to a large extent. Note that these actions allows for the study on video-based methods in Section \ref{sec:experiments:the_unreasonable_effectiveness_of_data}.

\vspace{2mm}
\noindent\textbf{Locations.}
The conventional optical \cite{Sigal2009HumanEvaSV, ionescu2013human3} or multi-view motion capture systems \cite{Joo2019PanopticSA, Yu2020HUMBIAL} require indoor environments, resulting in the scarcity of in-the-wild backgrounds. Thanks to the open-world design of GTA, we have seamless access to various locations with diverse backgrounds, from city streets to the wilderness. Our investigation in Section \ref{sec:experiments:the_unreasonable_effectiveness_of_data} highlights these diverse locations are complementary to real datasets that are typically collected indoor.

\noindent\textbf{Camera Angle.}
Recent studies \cite{madan2020and, cai2020messytable} have shown the critical impact of camera angles on model performance, yet its effect in 3D human recovery is not fully explored due to data scarcity: it is common to have datasets with fixed camera positions \cite{Sigal2009HumanEvaSV, ionescu2013human3, Joo2019PanopticSA, mehta2017monocular}. In \name, we choose to sample random camera positions from the distribution of the real datasets \cite{ionescu2013human3, mehta2017monocular, von2018recovering} to balance both diversity and realness. Our data collection tool enables the control of camera placement position and orientation, thus allowing the study of camera angles that is otherwise difficult in real life. Note that the global orientation of SMPL annotation in the camera coordinates is used to compute an elevation-azimuth representation of the camera angles relative to the subject in the canonical coordinates. We visualize the camera angles in Fig. \ref{fig:cam_angles}.

\vspace{2mm}
\noindent\textbf{Interaction.}
Compared to existing works that crop and paste subjects onto random backgrounds \cite{Varol2017LearningFS, mehta2017monocular}, the subjects in \name are rendered together with the scenes to achieve a more realistic subject-environment interaction empowered by the physics engine. Interesting examples include the subject falling off the edge of a high platform, and the subject stepping into a muddy pond causing water splashing. Moreover, taking advantage of the occlusion culling mechanism \cite{fabbri2018learning}, we are able to annotate the body joints as ``visible" to the camera, ``occluded" by other objects, or ``self-occluded" by the subject's own body parts. 

\vspace{2mm}
\noindent\textbf{Lighting and Weather.}
Instead of adjusting image exposure to mimic different lighting, we directly control the in-game time to sample data around the clock. Consequently, \name contains drastically different lighting conditions and shadow projections. We also introduce random weather conditions such as rain and snow to the scenes that would be otherwise difficult to capture in real life. 
\section{Experiments}
\label{sec:experiments}

In this section, we study how to use game-playing data for 3D human recovery for real-life applications. 

\subsection{Experiment Details}
\noindent\textbf{Datasets.}
We follow the original training convention of our baseline methods \cite{kanazawa2018end, kolotouros2019learning, kocabas2021pare}. we define the ``Real" datasets used in the experiments to include Human3.6M \cite{ionescu2013human3} (with SMPL annotations via MoSh \cite{loper2014mosh}), MPI-INF-3DHP \cite{mehta2017monocular}, LSP \cite{johnson2010clustered}, LSP-Extended \cite{Johnson2011LearningEH}, MPII \cite{andriluka20142d} and COCO \cite{lin2014microsoft}. ``Real" datasets consist of approximately 300K frames. "Blended" datasets are formed by simply mixing \name data with the ``Real" data. 
Amongst the standard benchmarks, 3DPW \cite{von2018recovering} has 60 sequences (51k frames) of unconstrained scenes. In contrast, MPI-INF-3DHP \cite{mehta2017monocular} has only two sequences of real outdoor scenes (728 frames) and Human3.6M \cite{ionescu2013human3} is fully indoor. Hence, we follow the convention \cite{kolotouros2019learning, kocabas2020vibe, kocabas2021pare} to evaluate models mainly on 3DPW test set to gauge their in-the-wild performances. Nevertheless, we also provide experiment results on Human3.6M and MPI-INF-3DHP.

\vspace{2mm}
\noindent\textbf{Metrics.}
The standard metrics are Mean Per Joint Position Error (MPJPE), and Procrustes-aligned \cite{gower1975generalized} Mean Per Joint Position Error (PA-MPJPE), \ie, MPJPE evaluated after rigid alignment of the predicted and the ground truth joint keypoints, both in millimeters ($mm$). We highlight that PA-MPJPE is the \textit{primary} metric \cite{kolotouros2019learning, Joo2020ExemplarFF}, on which we conduct most of our discussions. 

\begin{table*}[t]
  \centering
  \caption{\textbf{\name's impact on model performance}. The values are reported on 3DPW test set in mm. We employ two strategies: \textbf{blended training (BT)} that directly mixes \name data with real data to train an HMR model; \textbf{finetuing (FT)} that finetunes pretrained models with mixed data. Significant performance improvements are achieved with both settings. Including \name in the training boosts the HMR \cite{kanazawa2018end} baseline to outperform much more sophisticated methods such as SPIN \cite{kolotouros2019learning} that leverages in-the-loop optimization (Registration) and VIBE \cite{kocabas2020vibe} that utilizes temporal information (Video); State-of-the-art method PARE \cite{kocabas2021pare} also benefit from data mixture. We also conduct further experiments on video-based human recovery with VIBE in Table \ref{tab:effectiveness_on_video}. 
  Mixture: data mixture strategies. Real: real datasets. 
  }
  \label{tab:effectiveness_on_image}
  \begin{tabular}{lccclcccc}
    \hline
    Method & Mixture & Registration & Video & Pretrain & Train & Finetune & 
    MPJPE $\downarrow$ & PA-MPJPE $\downarrow$ \\
    \hline
    HMR & - & -  & -  & ImageNet & Real & - &  112.3 & 67.5 \\
    HMR+ & - & -  & -  & ImageNet & Real & - &  98.5   & 61.7 \\
    SPIN & - & \checkmark & - & ImageNet & Real & - &  96.9   & 59.2 \\
    VIBE  & - & - & \checkmark & ImageNet & Real & - &  93.5   & 56.5 \\ 
    PARE & - & - & - & ImageNet & Real & - & 82.0 & 50.9 \\
    \hline
    HMR & BT & - & - & ImageNet & Mixed &  - 
    & 98.7 \textcolor{ao}{(-13.6)} & 60.5 \textcolor{ao}{(-7.0)} \\
    HMR & FT & - & - & HMR & - & Mixed
    & 91.4 \textcolor{ao}{(-20.9)} & 55.7 \textcolor{ao}{(-11.8)} \\
    HMR+ & BT & - & - & ImageNet & Mixed &  - 
    & 88.7 \textcolor{ao}{(-9.8)} & 56.0 \textcolor{ao}{(-5.7)} \\
    HMR+ & FT  & - & - & HMR+ & - & Mixed & 
    91.3 \textcolor{ao}{(-7.2)} & 55.5 \textcolor{ao}{(-6.2)} \\
    SPIN & FT  & - & - & SPIN  & - & Mixed & 
    83.2 \textcolor{ao}{(-13.7)} & 52.0 \textcolor{ao}{(-7.2)} \\
    PARE & FT  & - & - & PARE  & - & Mixed & 
    77.5 \textcolor{ao}{(-4.5)} & 46.8 \textcolor{ao}{(-4.1)} \\
    \hline
  \end{tabular}
\end{table*}

\vspace{2mm}
\noindent\textbf{Training Details.}
We follow the original paper in implementing baselines \cite{kanazawa2018end, kolotouros2019learning, kocabas2020vibe, kocabas2021pare} on the PyTorch-based framework MMHuman3D \cite{mmhuman3d}. HMR+ is a stronger variant of the original HMR, for which we remove all adversarial modules from the original HMR \cite{kanazawa2018end} for fast training and add pseudo SMPL initialization (``static fits") for keypoint-only datasets following SPIN \cite{kolotouros2019learning} without further in-the-loop optimization. For the Blended Training (BT), since \name has a much larger scale than existing datasets, we run all our experiments on 32 V100 GPUs, with the batch size of 2048 (four times as SPIN \cite{kolotouros2019learning}). The learning rate is also scaled linearly by four times to 0.0002. The rest of the hyperparameters are the same as SPIN \cite{kolotouros2019learning}. For the Finetuning (FT) experiments, we use the learning rate of 0.00001 with the batch size of 512, on 8 V100 GPUs for two epochs.

\vspace{2mm}
\noindent\textbf{Domain Adaptation Training Details.}
We use the same training settings as Blended Training, except that an additional domain adaptation loss is added in training. CycleGAN \cite{zhu2017unpaired}, we first train a CycleGAN between real data and our synthetic \name data. Then we use a trained sim2real generator from the CycleGAN to transform the input \name image into a real-style image during training. For JAN~\cite{long2017deep}, we use the default Gaussian kernel with a bandwidth 0.92, and set its loss weight to 0.001. For Chen \etal~\cite{chen2021representation}, we use the default trade-off coefficient 0.1, and set its loss weight to 1e-4. For Ganin \etal \cite{ganin2015unsupervised}, we use a 3-layer MLP to classify the domain of given features extracted from the backbone. The loss weight of the adversarial part is progressively increased to 0.1 for more stable training.

\begin{table}[t]
  \centering
  \caption{\textbf{Video-based 3D human recovery.} The values are reported on 3DPW \cite{von2018recovering} test set with VIBE as the base model. MI3: MPI-INF-3DHP. GTA: \name. PA: PA-MPJPE. Accel: acceleration error ($mm/s^2$). *: downsampled \name data to match the size of MPI-INF-3DHP (96K SMPL poses).}
  \label{tab:effectiveness_on_video}
  \begin{tabular}{cccccc}
    \hline
     MI3 & 3DPW & \name & MPJPE $\downarrow$ & PA $\downarrow$ & Accel $\downarrow$ \\
    \hline
    \cm & -   & -           & 95.0 & 56.5 & 27.1           \\
    -   & \cm & -           & 87.9 & 54.7 & \textbf{23.2}  \\
    \cm & \cm & -           & 93.9 &  55.9 & 27.0 \\
    -   & -   & \checkmark* & 93.7 & 55.0 & 26.3           \\
    -   & -   & \cm         & 91.3 & 54.1 & 24.7           \\
    -   & \cm & \cm         & \textbf{85.2} & 52.4 & 24.2  \\
    \cm & \cm & \cm         & 86.0 & \textbf{51.9} & 23.3  \\
    \hline
  \end{tabular}
\end{table}

\subsection{Better 3D Human Recovery with Data Mixture}
\label{sec:experiments:the_unreasonable_effectiveness_of_data}

Despite that \name features reasonably realistic data, there inevitably exists domain gaps. Surprisingly, intuitive methods of data mixture are effective despite the domain gaps for both image- and video-based 3D human recovery.

\vspace{2mm}
\noindent\textbf{Image-based 3D Human Recovery.}
We evaluate the use of synthetic data under two data mixture settings: blended training (BT) and finetuning (FT). Results are collated in Table \ref{tab:effectiveness_on_image}. In \textbf{blended training (BT)}, synthetic \name data is directly mixed with a standard basket of real datasets \cite{kolotouros2019learning} (Human3.6M \cite{ionescu2013human3}, MPI-INF-3DHP \cite{mehta2017monocular}, LSP \cite{johnson2010clustered}, LSP-Extended \cite{Johnson2011LearningEH}, MPII \cite{andriluka20142d} and COCO \cite{lin2014microsoft}). Compared with the HMR and HMR+ baselines, blended training achieves 7.0 mm and 5.7 mm improvements in PA-MPJPE, surpassing methods such as SPIN \cite{kolotouros2019learning} that requires online registration or VIBE~\cite{kocabas2020vibe} that leverages temporal information. 
As for \textbf{finetuning (FT)}, we finetune a pretrained model with mixed data. Since finetuing is much faster than blended training, this allows us to perform data mixture on more base methods such as SPIN \cite{kolotouros2019learning} and PARE \cite{kocabas2021pare}. Finetuning leads to considerable improvements in PA-MPJPE compared to the original HMR (11.8 mm), HMR+ (6.2 mm), SPIN (7.2 mm) and PARE (4.1 mm) baselines. 

\vspace{2mm}
\noindent\textbf{Video-based 3D Human Recovery.}
In Table \ref{tab:effectiveness_on_video}, we validate that data mixture is also effective for video-based methods. We conduct the study with the popular VIBE \cite{kocabas2020vibe} as the base model. VIBE uses a pretrained SPIN model as the feature extractor for each frame, and we train the temporal modules with datasets indicated in Table \ref{tab:effectiveness_on_video}. The results of training with MI3 and 3DPW are quoted from the official codebase~\footnote{\url{https://github.com/mkocabas/VIBE/issues/99\#issuecomment-708351802}}. We obtain the following observations. First, when training alone, \name outperforms MPI-INF-3DHP with an equal number of training data. Second, the full set of \name is comparable with the in-domain training source (3DPW train set), even slightly better in PA-MPJPE. Third, \name is complementary to real datasets as blended training leads to highly competitive results in all metrics.

\vspace{2mm}
\noindent\textbf{Comparison with Other Data-driven Methods.}
We highlight that \name is a large-scale, diverse dataset for 3D human recovery. In Table \ref{tab:other_data_driven}, we compare \name with several other recent works that provide additional data for human pose and shape estimation. We show that \name is a practical training source that improves the performance of various base methods. Notably, \name slightly surpasses AGORA, which is built with expensive industry-level human scans of high-quality geometry and texture. This result suggests that scaling with game-playing data at a lower cost achieves a similar effect.
\begin{table}[t]
  \centering
  \caption{\textbf{Comparison with other data-driven methods.} \name data effectively improves the base method performance. The numbers are reported on 3DPW test set, \textit{without} using 3DPW in the training. *Blended AGORA and real data for a fair comparison.
  }
  \label{tab:other_data_driven}
    \begin{tabular}{lcrcc}
    \hline
    Dataset & Method & MPJPE $\downarrow$ & PA-MPJPE
    $\downarrow$ \\
    \hline
    Arnab \etal \cite{arnab2019exploiting} & HMR
    & - & 72.2 \\
    
    EFT \cite{Joo2020ExemplarFF} & SPIN
    & - & 54.2 \\
    
    AGORA \cite{Patel:CVPR:2021} & SPIN
    & 85.7 & 55.3 \\

    AGORA* \cite{Patel:CVPR:2021} & SPIN & 
    84.4 & 54.9  \\

    \hline
    \name (BT) & HMR
    & 98.7 & 60.5 \\
    
    \name (FT) & HMR
    & 91.4 & 55.5 \\
    
    \name (FT) & SPIN
    & \textbf{83.1} & \textbf{52.0} \\
    
    \hline
  \end{tabular}
\end{table}

\subsection{Closing the Domain Gap with Synthetic Data}
\label{sec:experiments:closing_the_gap}
After obtaining good results under both image- and video-based settings on 3DPW, an in-the-wild dataset and the standard test benchmark, we extend our study to answer \textit{why is game playing data effective at all?} \textcolor{blue}{We highlight that this investigation holds more significance to the community as it provides fundamental reasons for the use of synthetic data.} To this end, we also evaluate models on other (mostly) indoor benchmarks such as Human3.6M (Protocol 2) and MPI-INF-3DHP in Table \ref{tab:more_benchmarks}. Interestingly, we notice that the performance gains on these two benchmarks are not as significant as those on 3DPW. Moreover, existing methods \cite{kanazawa2018end, kolotouros2019learning} commonly include in-the-wild COCO data in the training set, in addition to popular training datasets are typically collected indoors. We aim to explain the above-mentioned observations and practices through both qualitative and quantitative evaluations. 

\begin{figure}[t]
  \centering
  \vspace{-2mm}
  \includegraphics[width=\linewidth]{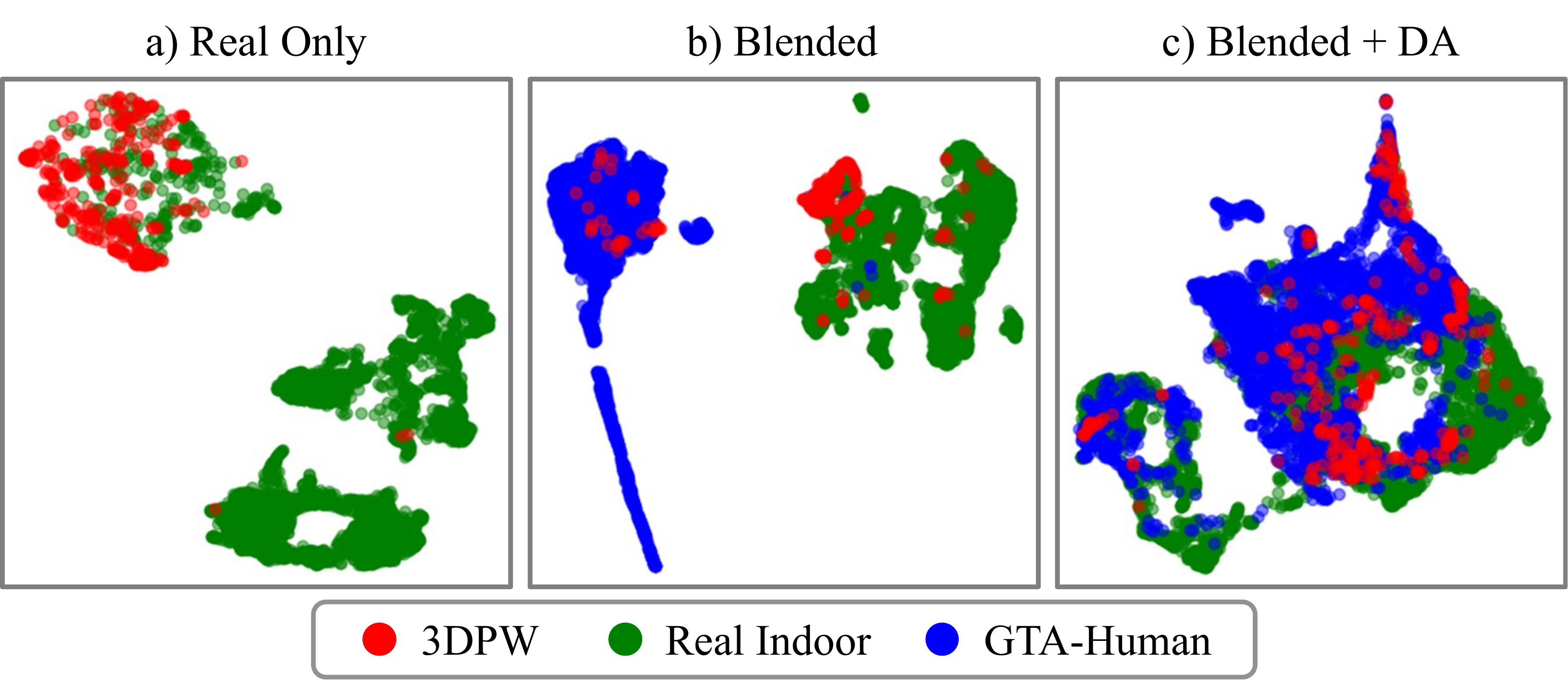}
  \vspace{-5mm}
  \caption{\textbf{Domain gap analysis.} We visualize features extracted after the trained backbones via UMAP \cite{mcinnes2018umap} dimension reduction (the two axes are the principal axes). (a) Training with real data only. (b) Blended training. (c) Blended training with domain adaptation (Ganin \etal \cite{ganin2015unsupervised}).}
  \label{fig:domain_gap_analysis}
\end{figure}

In Fig. \ref{fig:domain_gap_analysis}, we visualize the feature distribution of various datasets. We discover that there are indeed some domain gaps between real indoor data and real outdoor data. Hence, models trained on real indoor data may not perform well in the wild. We observe that in Fig. \ref{fig:domain_gap_analysis}(a), indoor data has a significant domain shift away from in-the-wild data. This result implies that models trained on indoor datasets may not transfer well to in-the-wild scenes. In Fig.\ref{fig:domain_gap_analysis}(b), blended training achieves better results as 3DPW test data are well-covered by mixing real data or \name data. Specifically, even though the domain gap between \name and real datasets persists, the distribution of 3DPW data is split into two main clusters, covered by \name and real datasets separately. \textcolor{blue}{Hence, this novel observation may explain the effectiveness of \name: albeit synthetic, a large amount of in-the-wild data provides meaningful knowledge that is complementary to the real datasets.}

Moreover, we further validate the synergy between real and synthetic data through domain adaptation in Table \ref{tab:domain_adaptation}. We select and implement several mainstream domain adaption methods \cite{wiles2021fine}, and evaluate them on an HMR model under BT with an equal amount of real data and \name ($1\times$). We discover that learned data augmentation such as CycleGAN \cite{zhu2017unpaired} may not be effective, whereas domain generalization techniques (JAN \cite{long2017deep} and Ganin \etal \cite{ganin2015unsupervised}) and domain adaptive regression such as Chen \etal \cite{chen2021representation} further improves the performance. In Fig.\ref{fig:domain_gap_analysis}(c), domain adaptation (Ganin \etal \cite{ganin2015unsupervised}) pulls the distributions of both real and \name data together, and they jointly establish a better-learned distribution to match that of the in-the-wild 3DPW data.

\begin{table}[t]
  \centering
  \caption{\textbf{More benchmarks}. We evaluate image-based methods trained with data mixture strategies on Human3.6M and MPI-INF-3DHP. We observe that the performance boosts are smaller than that on 3DPW. This may be attributed to the indoor-outdoor domain gaps that we discuss in Section \ref{sec:experiments:closing_the_gap}.}
  \label{tab:more_benchmarks}
  \begin{tabular}{lccccc}
    \hline
    & & \multicolumn{2}{c}{Human3.6M} & \multicolumn{2}{c}{MPI-INF-3DHP} \\
        \cmidrule(lr){3-4} \cmidrule(lr){5-6}
    Method & Mixture & 
    MPJPE $\downarrow$ & PA $\downarrow$ & 
    MPJPE $\downarrow$ & PA $\downarrow$ \\
    \hline
    
    HMR  
    & - & 77.9 & 55.8 & 107.2 & 74.1  \\
    
    SPIN 
    & - & - & 41.1 & 105.2 & 67.5 \\
    
    
    \hline
    HMR 
    & BT & 74.5 & 51.3 & 103.4 & 71.3 \\
    
    HMR 
    & FT & 73.2 & 52.5 & 102.9 & 71.0 \\
    
    SPIN 
    & FT & 60.9 & 40.8 & 96.4 & 67.0 \\
    
    
    \hline
  \end{tabular}
  \vspace{-2mm}
\end{table}
\begin{table}[t]
  \caption{\textbf{Domain adaptation} with equal amount real and synthetic data. PA: PA-MPJPE.}
  \label{tab:domain_adaptation}
  \centering
  \begin{tabular}{lccc}
    \hline
    Method & Real & \name & PA-MPJPE $\downarrow$ \\
    \hline
    HMR & \checkmark & - & 76.7 \\
    HMR ($1\times$) & - & \checkmark & 65.7 \\
    \hline
    HMR (BT, $1\times$) & \checkmark & \checkmark & 58.6\\
    \hline
    CycleGAN \cite{zhu2017unpaired} &  \checkmark & \checkmark & 61.6 \\
    Chen \etal \cite{chen2021representation} &  \checkmark & \checkmark & 57.9\\
    JAN \cite{long2017deep} &  \checkmark & \checkmark & 56.5 \\
    Ganin \etal \cite{ganin2015unsupervised} &  \checkmark & \checkmark & \textbf{55.5} \\
    \hline
  \end{tabular}
\end{table}

\begin{figure}[t]
  \centering
  \includegraphics[width=\linewidth]{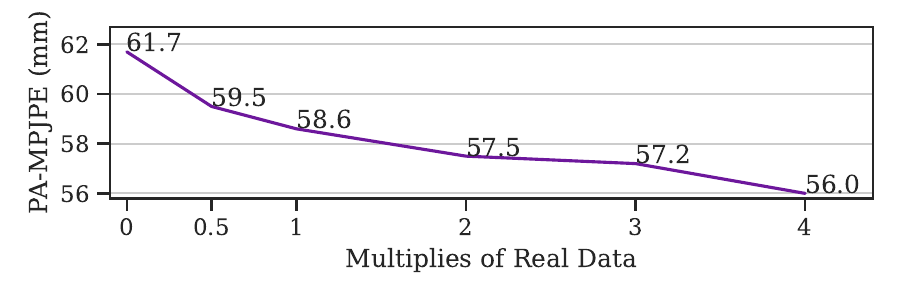}
  \vspace{-6mm}
  \caption{\textbf{Amount of \name Data.} The horizontal axis indicate the amount of \name data used as multiples of the amount of real data. HMR+ is used as the base method.}
  \label{fig:data_amount}
\end{figure}

\subsection{Dataset Scale Matters}
\label{sec:experiments:more_data_is_better}

We study the data scale in two aspects. 1) Different amounts of \name data are progressively added in the training to observe the trend in the model performance. 2) The influence of a lack of data from the perspectives of critical factors such as camera angle, pose, and occlusion.

\vspace{2mm}
\noindent\textbf{Amount of \name Data.}
In Fig. \ref{fig:data_amount}, we delve deeper into the impact of data quantity on 3DPW test set. HMR is used as the base model with BT setting. The amount of \name data used is expressed as multiples of the total quantity of real datasets ($\sim$300K \cite{kolotouros2019learning}). For example, $2\times$ means the amount of \name is twice as much as the real data in the BT. A consistent downward trend in the errors ($\sim$6 mm decrease) with increasing \name data used in the training is observed. Since real data is expensive to acquire, synthetic data may play an important role in scaling up 3D human recovery in real life.

\vspace{2mm}
\noindent\textbf{Synthetic Data as a Scalable Supplement.}
We collate more experiments with different real-synthetic data ratios in Table \ref{tab:data_ratio}, using HMR+ as the base method and BT as the data mixture strategy. We observe that 1) Adding more data, synthetic and real alike, generally improve the performance. 2) Mixing 75\% real data with 25\% synthetic data performs well (200K to 400K data). 3) When the data amount increases, high ratio of real data cannot be sustained beyond 300K data due to insufficient real data. However, additional synthetic data still improves model performance. These experiments reaffirm that synthetic data complements real data. \textcolor{blue}{More importantly, in real practice, synthetic data can serve as an easily scalable training source to supplement typically limited real data that is too expensive to accumulate further.}

\begin{figure*}[t]
  \centering
  \includegraphics[width=\linewidth]{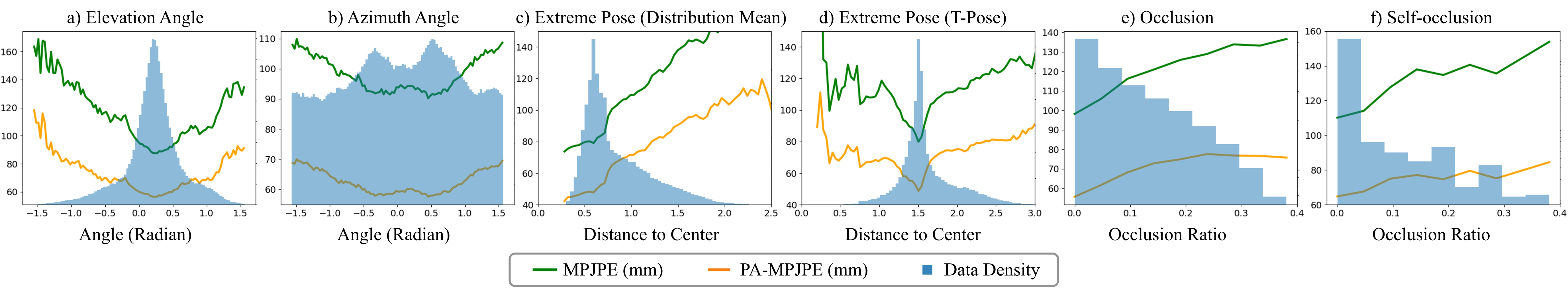}
  \vspace{-5mm}
  \caption{\textbf{Impact of data scarcity.} We show that model performance is sensitive to data scarcity, and this observation is consistent on factors such as camera angles, poses, and occlusion. For c) and d), we follow \cite{rong2020chasing} to encode pose as a set of 3D coordinates of the 24 key joints, and plot the distance from the mean pose and T-pose respectively. The data density of e) and f) are in log scale.}
  \label{fig:systematic_study}
\end{figure*}

\vspace{2mm}
\noindent\textbf{Impact of Data Scarcity.}
In Fig. \ref{fig:systematic_study}, we systematically study the HMR+ model trained with BT and evaluate its performance on \name, subjected to different data density for factors such as camera angle, pose, and occlusion. We discretize all examples evaluated to obtain and plot the data density with bins, and compute the mean error for each bin to form the curves. 
A consistent observation across factors is that the model performance deteriorates drastically when data density declines, indicating high model sensitivity to data scarcity. Hence, strategically collected synthetic data may effectively supplement the real counterpart, which is often difficult to obtain. 

\begin{table}[t]
  \centering
  \caption{\textbf{Synthetic Data as a Supplement.} Different total data amount with different real data ratio are shown. Values are PA-MPJPE (mm) on 3DPW test set. Synthetic data are sampled from 4$\times$ set during training. N/A: this ratio cannot be sustained beyond 300K data due to insufficient real data. HMR+ (BT) is used as the base method.}
  \label{tab:data_ratio}
  \begin{tabular}{lccccc}
    \hline
    Real Ratio  & 100K & 200K & 300K & 400K & 500K  \\
    \hline
    0\%   & 70.6 & 64.5 & 65.7 & 65.0 &	64.9 \\
    25\%  & 62.4 & 60.9 & 58.0 & 57.6 & 57.3 \\
    50\%  & 61.7 & 58.9 & 57.9 & 56.3 & 55.6 \\
    75\%  & 62.4 & 58.4 & 56.8 & 55.7 & N/A \\
    100\% & 65.8 & 62.7 & 61.7 & N/A & N/A \\
    \hline
  \end{tabular}
\end{table}
\begin{table}[t]
  \centering
  \caption{\textbf{Strong supervision is key}. The first row is the HMR+ baseline without any \name data added.}
  \label{tab:strong_supervision_is_key}
  \begin{tabular}{ccccc}
    \hline
    Keypoints & SMPL & MPJPE $\downarrow$ & PA-MPJPE $\downarrow$ \\
    \hline
    -          & -            & 98.5 & 61.7 \\
    \checkmark & -           & 93.4 & 60.9 \\
    -          & \checkmark & 92.0 & 56.3\\
    \checkmark & \checkmark & 88.7 & 56.0\\
    \hline
  \end{tabular}
\end{table}

\subsection{Strong Supervision is Key}
\label{sec:experiments:strong_supervision_is_key}
Due to the prohibitive cost of collecting a large amount of SMPL annotations with a real setup, it is appealing to generate synthetic data that is automatically labelled. In this section, we investigate the importance of strong supervision and discuss the reasons. 
We compare weak supervision signals (\ie, 2D and 3D keypoints) to strong counterparts (\ie, SMPL parameters), and find out that the latter is critical to training a high-performing model. We experiment under BT setting on 3DPW test set, Table \ref{tab:strong_supervision_is_key} shows that strong supervision of SMPL parameters $\theta$ and $\beta$, are much more effective than weak supervision of body keypoints. 
Our findings are in line with SPIN \cite{kolotouros2019learning}. SPIN tests fitting 3D SMPL on 2D keypoints to produce pseudo SMPL annotations during training and finds this strategy effective. However, this conclusion still leaves the root cause of the effectiveness of 3D SMPL unanswered, as recent work suggests that 2D supervision is inherently ambiguous \cite{kissos2020beyond}. \textcolor{blue}{Note that in this work, we extend the prior study on the supervision types by adding in 3D keypoints as a better part of the weak supervision and find out that SMPL annotation is still far more effective.}

\textcolor{blue}{More importantly, we wish to offer preliminary discussions on what makes strong supervision (SMPL parameters) more effective than weak supervision (keypoints).} We argue that keypoints only provide partial guidance to body shape estimation $\beta$ (bone length only), but $\beta$ is required in joint regression from the parametric model. Moreover, ground truth SMPL parameters is directly used in the loss computation with the predicted SMPL parameters (Equation \ref{eq:l_smpl}), which initiates gradient flow that reaches the learnable SMPL parameters in the shortest possible route. On the contrary, the 3D keypoints $\hat{X}_{3D}$ are obtained with joint regression $\mathcal{J}$ of canonical keypoints with estimated body shape $\hat{\beta}$, and the global rigid transformation $M$ derived from the SMPL kinematic tree (Equation \ref{eq:l_3d}). The 2D keypoints $\hat{X}_{2D}$ further require extra estimation of translation $\mathbf{\hat{t}}$ for the transformation $\mathcal{T}$ of the 3D keypoints, and 3D to 2D projection $\mathcal{K}$ with assumed focal length $\mathbf{f}$ as well as camera center $\mathbf{c}$. The elongated route and uncertainties introduced in the process to compute the loss for 2D keypoints (Equation \ref{eq:l_2d}) hinder the effective learning.
\begin{align}
\mathcal{L}_{SMPL}&=||\theta-\hat{\theta}||+||\beta-\hat{\beta}|| 
\label{eq:l_smpl} \\
\mathcal{L}_{3D}&=||\hat{X}_{3D}-X_{3D}||  
\label{eq:l_3d} \\
\mathcal{L}_{2D}&=||\hat{X}_{2D}-X_{2D}||
\label{eq:l_2d}
\end{align} 
where 
\begin{align}
\hat{X}_{3D}&=\mathcal{M}(J(\hat{\beta}), \hat{\theta}) \\
\hat{X}_{2D}&=\mathcal{K}(\mathcal{T}(\hat{X}_{3D}, \mathbf{\hat{t}}), \mathbf{f}, \mathbf{c})
\end{align}

\subsection{Big Data Benefits Big Models}
\label{sec:experiments:big_data_and_big_model}
ResNet-50 remains a common backbone choice, since HMR is firstly introduced for deep learning-based 3D human recovery. In this section, we extend our study of the impact of big data on more backbone options, including deeper CNNs such as ResNet-101 and 152 \cite{he2016deep}, as well as DeiT \cite{touvron2020training}, as a representative of Vision Transformers. 
In Table \ref{tab:big_data_and_big_model}, we evaluate various backbones for the HMR baseline. We highlight that including \name always improves model performance by a considerable margin, regardless of the model size or architecture.
Note that using Transformers as the feature extractor for human pose and shape estimation is under-explored in recent literature; there may be some room for further improvement upon our attempts presented here. Nevertheless, the same trend holds for the two transformer variants.
Interestingly, additional \name unleashes the full power of a small model (\eg, ResNet-50), enabling it to outperform a larger model (\eg, ResNet-152) trained with real data only. This suggests data still remains a critical bottleneck for accurate human pose and shape estimation.

\begin{table}[t]
  \centering
  \caption{\textbf{Big data benefits big models.} Real: training with only the real datasets. +GTA: blended training setting is used with \name. Values in green indicate the error reduction in PA-MPJPE (mm) with blended training.}
  \label{tab:big_data_and_big_model}
  \begin{tabular}{lccc}
    \hline
    Backbone    & \#Param & Real $\downarrow$ & +\name $\downarrow$\\
    \hline
    ResNet-50   & 26M & 61.7 &  56.0 \textcolor{ao}{(-5.7)}\\
    ResNet-101  & 45M & 60.1 &  54.5 \textcolor{ao}{(-5.6)}\\
    ResNet-152  & 60M & 58.4 &  54.3 \textcolor{ao}{(-4.1)}\\
    \hline
    DeiT-Small  & 22M & 66.5 & 60.7 \textcolor{ao}{(-5.8)}\\
    DeiT-Base   & 86M & 61.2 & 56.2 \textcolor{ao}{(-5.0)}\\
    \hline
  \end{tabular}
\end{table}

\section{Conclusion}
In this work, we evaluate the effectiveness of synthetic game-playing data in enhancing human pose and shape estimation especially in the wild. To this end, we present \name, a large-scale, diverse dataset for 3D human recovery. Our experiments on \name provide five takeaways: 
\textbf{1)} Training with diverse synthetic data (especially with outdoor scenes) achieves a significant performance boost.
\textbf{2)} The effectiveness is attributed to the complementary relation between real and synthetic data.
\textbf{3)} The more data, the better because model performance is highly sensitive to data density.
\textbf{4)} Strong supervision such as SMPL parameters are essential to training a high-performance model. 
\textbf{5)} Deeper and more powerful backbones also benefit from a large amount of data. 
As for future works, we plan to investigate beyond 1.4M data samples with more computation budgets to explore the boundary of training with synthetic data. Moreover, it would be interesting to study the sim2real problem for 3D parametric human recovery more in-depth with \name, or even extend the game-playing data to other human-related topics such as model-free reconstruction that are out of the scope of this work. In addition, the annotation pipeline may be upgraded to fully capture the body shape besides the bone lengths from the 3D mesh of the subjects.

\ifCLASSOPTIONcompsoc
  \section*{Acknowledgments}
\else
  \section*{Acknowledgment}
\fi

This study is supported by the Ministry of Education, Singapore, under its MOE AcRF Tier 2 (MOET2EP20221-0012), NTU NAP, and under the RIE2020 Industry Alignment Fund – Industry Collaboration Projects (IAF-ICP) Funding Initiative, as well as cash and in-kind contribution from the industry partner(s).

\appendices

\section{Ethic Concerns}

As we are aware that GTA is not a perfect depiction of real life, we address some ethical concerns and explain our strategies to alleviate potential negative impact.

\subsection{Privacy}

The subjects present in GTA-Human are virtual humans extracted from the in-game database, that usually do not have clear real-life references. Protagonists may have some sort of real-life references, but the appearances are altered to suit the corresponding characters in the context of the game storyline. 

\subsection{Violence and sexualized actions}

We manually screen around 1k actions from the 20k classes and find that the vast majority of the actions are used to depict the ordinary lives of the city-dwellers (e.g. walking, drinking coffee, doing push-ups, and so on). This is further supported by 1) the distribution of GTA-Human is center-aligned with real datasets. 2) methods trained on GTA-Human can perform convincingly better on standard real datasets. Both indicate that the domain shift in actions may not be noticeably affected by the small portion of offensive actions. Moreover, no weapon is depicted in the dataset.

\subsection{Stereotypes and Biases}

The original storylines of GTA-V may insert strong stereotypes in the depiction of characters depending on their attributes (\eg, gender and skin tones). We thus adopt the following strategies to minimize biases.

First, all factors including the characters and actions, are decoupled and randomized in \name. Specifically, the examples in \name are not linked to the original storylines; all the characters and the actions are pulled out of the in-game database, randomly assigned at random locations all over the map. Hence, it is very unlikely any character-specific actions could be reproduced. 

Second, we have conducted a manual analysis on \textcolor{blue}{skin tone and clothing. Our inspection shows the database consists of 40\% with fair to light skin tones, 29\% with light to medium skin tones, 17\% with medium to tan skin tones, 11\% with dark to deep skin tones, and 3\% undetermined}. As for clothing, we find it difficult to determine if a specific attire has certain social implications without context (for example, skin-showing attire not be associated with sex workers as it is also common to find people in bikinis at the beach), we thus categorize all clothing into formal, semi-formal and casual. We observe that while there is approximately the same number of men and women in formal attire (11\% vs 9\%); more men in casual attires (\eg tank tops, topless) than women (25\% vs 11\%); subjects of various skin tone categories have approximately the same distribution of formal, semi-formal and casual ($\sim 2:5:3$). Hence, we find the character appearances mostly (albeit not perfectly) balanced across genders and skin tones.

Third, as much as we hope to perform a complete and thorough data screening and cleaning, we highlight it is not very practical to manually inspect all examples due to the sheer scale of the dataset. Hence, we anonymize the characters, actions, and locations such that they exist in the dataset to enrich the distribution, but cannot be retrieved for malicious uses. 

\subsection{Copyright}

The publisher of GTA-V allows for the use of game-generated materials provided that it is non-commercial, and no spoilers are distributed\footnote{Policy on mods. \url{http://tinyurl.com/yc8kq7vn}.}\footnote{Policy on copyrighted material. \url{http://tinyurl.com/pjfoqo5}.}.
Hence, we follow prior works that generate data on the GTA game engine \cite{richter2016playing, richter2017playing, krahenbuhl2018free, fabbri2018learning, cao2020long, hu2021sail} to make \name publically available.

\section{Additional Experiments and Discussions}
\begin{table*}[t]
  \centering
  \begin{tabular}{ccccccccc}
    \toprule
    \multicolumn{3}{c}{} & 
    \multicolumn{2}{c}{EHF~\cite{pavlakos2019expressive}} &
    \multicolumn{2}{c}{AGORA-Val~\cite{Patel:CVPR:2021}} &
    \multicolumn{2}{c}{EgoBody (EgoSet)~\cite{zhang2022egobody}} \\
    \cmidrule(lr){4-5} \cmidrule(lr){6-7} \cmidrule(lr){8-9}
    Base Model & \#Params & GTA-Human & 
    PA-PVE & PVE & PA-PVE & PVE & PA-PVE & PVE \\
    \midrule
     SMPLer-X-S & 32M & - & 
     61.9 & 86.1 & 74.2 & 179.9 & 60.8 & 124.5
     \\

     SMPLer-X-S & 32M & \checkmark & 
     52.3\textcolor{ao}{(-9.6)} & 
     79.7\textcolor{ao}{(-6.4)} & 
     69.6\textcolor{ao}{(-4.6)} & 
     161.3\textcolor{ao}{(-18.6)} & 
     56.7\textcolor{ao}{(-4.1)} & 
     109.3\textcolor{ao}{(-15.2)}
     \\ 
     
     SMPLer-X-H & 662M & - & 
     49.4 & 73.3 & 66.7 & 166.4 & 55.0 & 115.1
     \\ 
     
     SMPLer-X-H & 662M & \checkmark & 
     45.2\textcolor{ao}{(-4.2)} & 
     61.8\textcolor{ao}{(-11.5)} & 
     60.7\textcolor{ao}{(-6.0)} & 
     143.6\textcolor{ao}{(-22.8)} &
     48.0\textcolor{ao}{(-7.0)} & 
     98.1\textcolor{ao}{(-17.0)}
     \\ 
     
    \bottomrule
  \end{tabular}
  \caption{Training models of different sizes with GTA-Human. GTA-Human brings substantial performance improvements for both small and large models. Surprisingly, the large model sometimes benefits more from additional synthetic data. SMPLer-X-S represents uses ViT-S backbone and represents small models whereas SMPLer-X-H uses the ViT-H backbone and represents large models. }
  \label{tab:rebuttal_larger_model}
\end{table*}

\subsection{Qualitative Evaluations}
\begin{figure}[h]
  \centering
  \includegraphics[width=\linewidth]{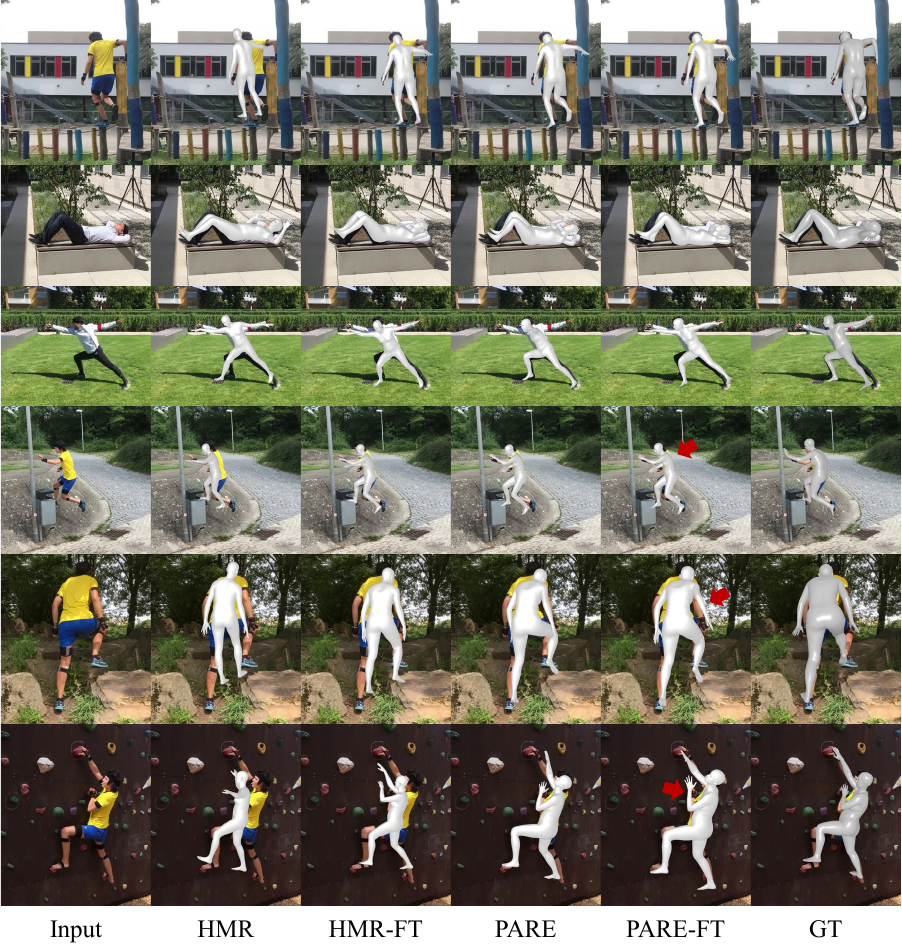}
  \caption{Qualitative results of various baseline methods on 3DPW. Despite that finetuning (FT) with GTA-Human significantly improves baseline methods, there is still room for improvement (indicated by red arrows).}
  \label{fig:additional_visual}
\end{figure}

We add a qualitative comparison between the models (HMR and PARE are used as the base methods) trained with (FT setting) and without GTA-Human in Fig.~\ref{fig:additional_visual}. It is observed that adding GTA-Human brings noticeable improvements in the estimations. However, there are still some failure cases (indicated by red arrows): a strong model trained with GTA-Human (PARE-FT) can still have some suboptimal estimations, resulting in misalignments between the parametric models and the subjects in the input image.

\subsection{\name and AGORA Are Complementary}
In this and the following sections, we extend GTA-Human to include SMPL-X annotations and use SMPLer-X~\cite{cai2023smpler} as the base method. Evaluation benchmarks are EHF~\cite{pavlakos2019expressive} and EgoBody (EgoSet)~\cite{zhang2022egobody}. We show in Table~\ref{tab:rebuttal_gta+agora} that GTA-Human and AGORA are complementary data sources. 
\begin{table}[t]
  \centering
  \begin{tabular}{cccccc}
    \toprule
    \multicolumn{2}{c}{} & 
    \multicolumn{2}{c}{EHF~\cite{pavlakos2019expressive}} &
    \multicolumn{2}{c}{EgoBody~\cite{zhang2022egobody}} \\
    \cmidrule(lr){3-4} \cmidrule(lr){5-6}
    AGORA & GTA-Human  & PA-PVE & PVE & PA-PVE & PVE \\
    \midrule
     \checkmark & - & 90.6 & 130.9 & 71.9 & 135.6 \\
     - & \checkmark & 55.3 & 120.0 &65.3 & 126.1 \\
     \checkmark & \checkmark & \textbf{54.3} & \textbf{91.9} & \textbf{58.7} & \textbf{113.6} \\
    \bottomrule
  \end{tabular}
  \caption{GTA-Human and AGORA are complementary data sources. GTA-Human is extended to provide SMPL-X annotation for a fair comparison with AGORA to train SMPLer-X as the base model. The results on EHF and EgoBody (EgoSet) show training with both GTA-Human and AGORA achieves the best results. }
  \label{tab:rebuttal_gta+agora}
\end{table}

\subsection{\name for Training Large Models}
In Table \ref{tab:rebuttal_larger_model}, we validate that GTA-Human is useful for large model training. More importantly, the large model (SMPLer-X-H) seems to benefit more from the addition of GTA-Human than the smaller counterpart for multiple metrics on various test benchmarks. 

\subsection{Case Study on Elevation Angles}
\begin{figure}[h]
  \centering
  \includegraphics[width=\linewidth]{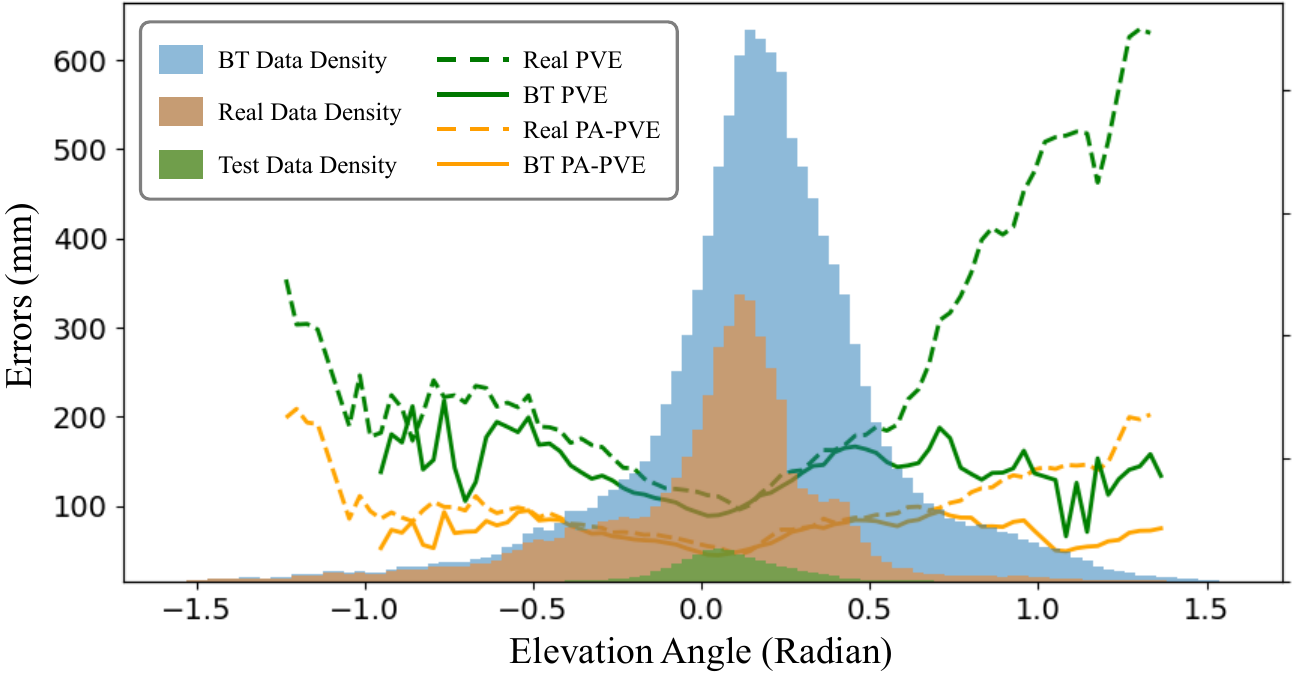}
  \caption{Blended training with GTA-Human improves method robustness on EgoBody with originally low data density at the right side of the figure. GTA-Human supplements the real data, especially in areas with large positive elevation angles (when cameras are looking down on the subject). Hence, blended training (BT) effectively mitigates the large errors. }
  \label{fig:additional_exp}
\end{figure}

We take the elevation angle as a case study and delve deeper into the performance comparison between "real" and "blended" training in Fig. ~\ref{fig:additional_exp}. Here we use SMPLer-X-S~\cite{cai2023smpler} as the base method, evaluated on EgoBody~\cite{zhang2022egobody} dataset. GTA-Human supplements real data with a considerable amount of data with diverse elevation angles, especially at the large positive region. As a result, the blended training (mixing real dataset and 100K GTA-Human data) results in a much more robust model at high elevation angles than the original one that is trained on real data only. 
.

\ifCLASSOPTIONcaptionsoff
  \newpage
\fi




{
\bibliographystyle{IEEEtran}
\bibliography{references}
}
\begin{IEEEbiography}[{\includegraphics[width=1in,height=1.25in,clip,keepaspectratio]{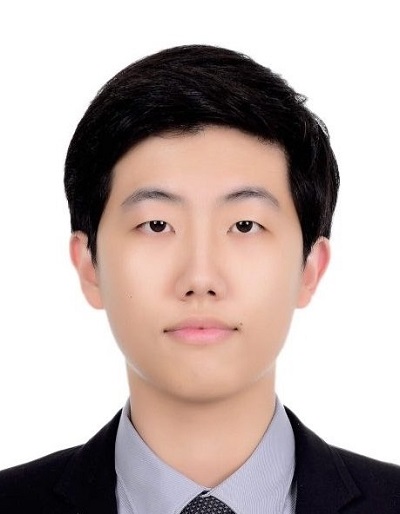}}]{Zhongang Cai}
  is currently a Ph.D. Student at Nanyang Technological University (NTU) advised by Prof. Ziwei Liu and Prof. Chen Change Loy, and a Senior Algorithm Researcher at SenseTime Research. He attained his bachelor’s degree at NTU and was awarded the Lee Kuan Yew Gold Medal as the top student. His research focuses on human motion capture and generation. To date, he has published 10+ papers on top venues such as ICLR, CVPR, ICCV, and ECCV. He is actively involved in academic services and was awarded the Outstanding/Highlighted Reviewer for ICLR'22 and ICCV'21.
\end{IEEEbiography}

\begin{IEEEbiography}[{\includegraphics[width=1in,height=1.25in,clip,keepaspectratio]{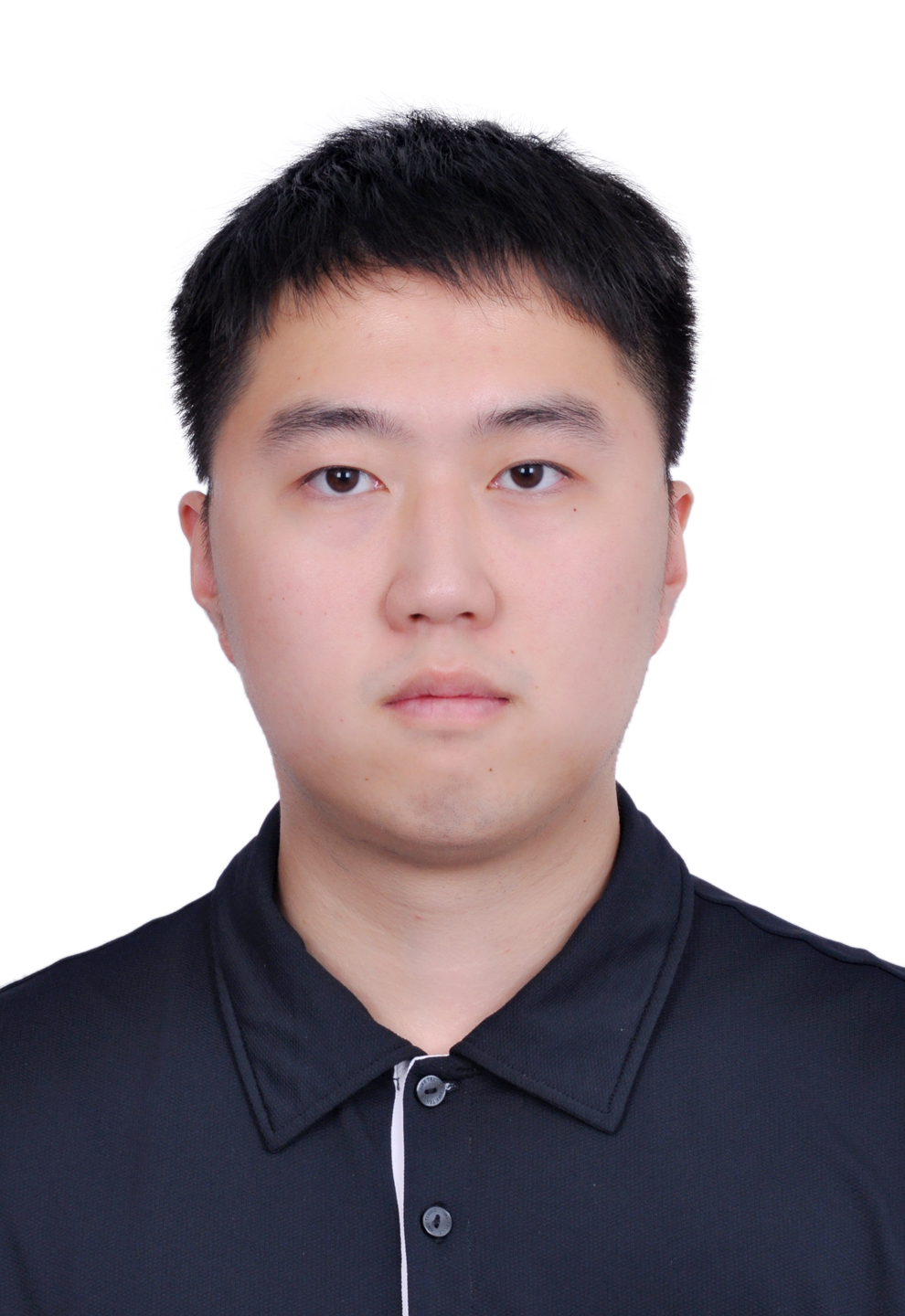}}]{Mingyuan Zhang}
  received a B.S. degree in computer science and engineering from Beihang University, China. He is currently pursuing a Ph.D. degree at MMLab@NTU, advised by Prof. Ziwei Liu. His research interests in computer vision include motion synthesis, 3D pose estimation, and scene understanding. He has published papers on top-tier conferences, including CVPR, ECCV, ICCV, ICLR, and AAAI.
\end{IEEEbiography}

\begin{IEEEbiography}[{\includegraphics[width=1in,height=1.25in,clip,keepaspectratio]{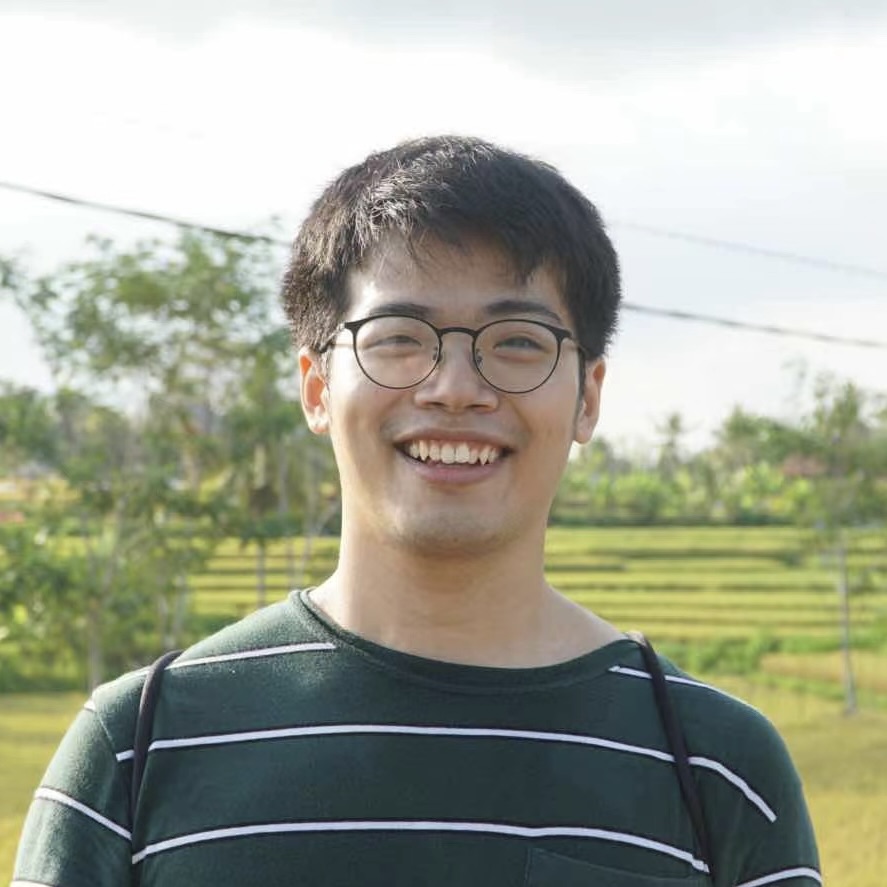}}]{Jiawei Ren}
  Jiawei Ren is a PhD student in Computer Science at MMLab@NTU, advised by Prof. Ziwei Liu. His research focuses on robust learning for 3D perception and generation. Previously, he was a full-time algorithm researcher at X-Lab@SenseTime Research where he was advised by Prof. Hongsheng Li. He obtained B.Eng. in Electrical and Electronic Engineering from Nanyang Technological University where he was advised by Prof. Lap-Pui Chau. He has published several papers on top-tier conferences including CVPR, ECCV, NeurIPS, ICML, and AAAI. He is also the recipient of AISG fellowship.
\end{IEEEbiography}

\begin{IEEEbiography}[{\includegraphics[width=1in,height=1.25in,clip,keepaspectratio]{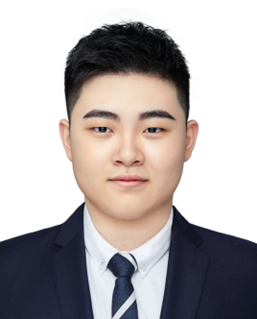}}]{Chen Wei}
  graduated from Nanyang Technological University with a bachelor's degree in Electric and Electronic Engineering. He was an algorithm research intern at S-Lab @ NTU in 2021 and joined SenseTime International Pte. Ltd in 2022. His work focuses on the generation of synthetic human data from game engines and the standardization of human-centric computer vision datasets. He has published several papers on top-tier conferences, including ICCV, NeurIPS, and CVPR. 
\end{IEEEbiography}

\begin{IEEEbiography}[{\includegraphics[width=1in,height=1.25in,clip,keepaspectratio]{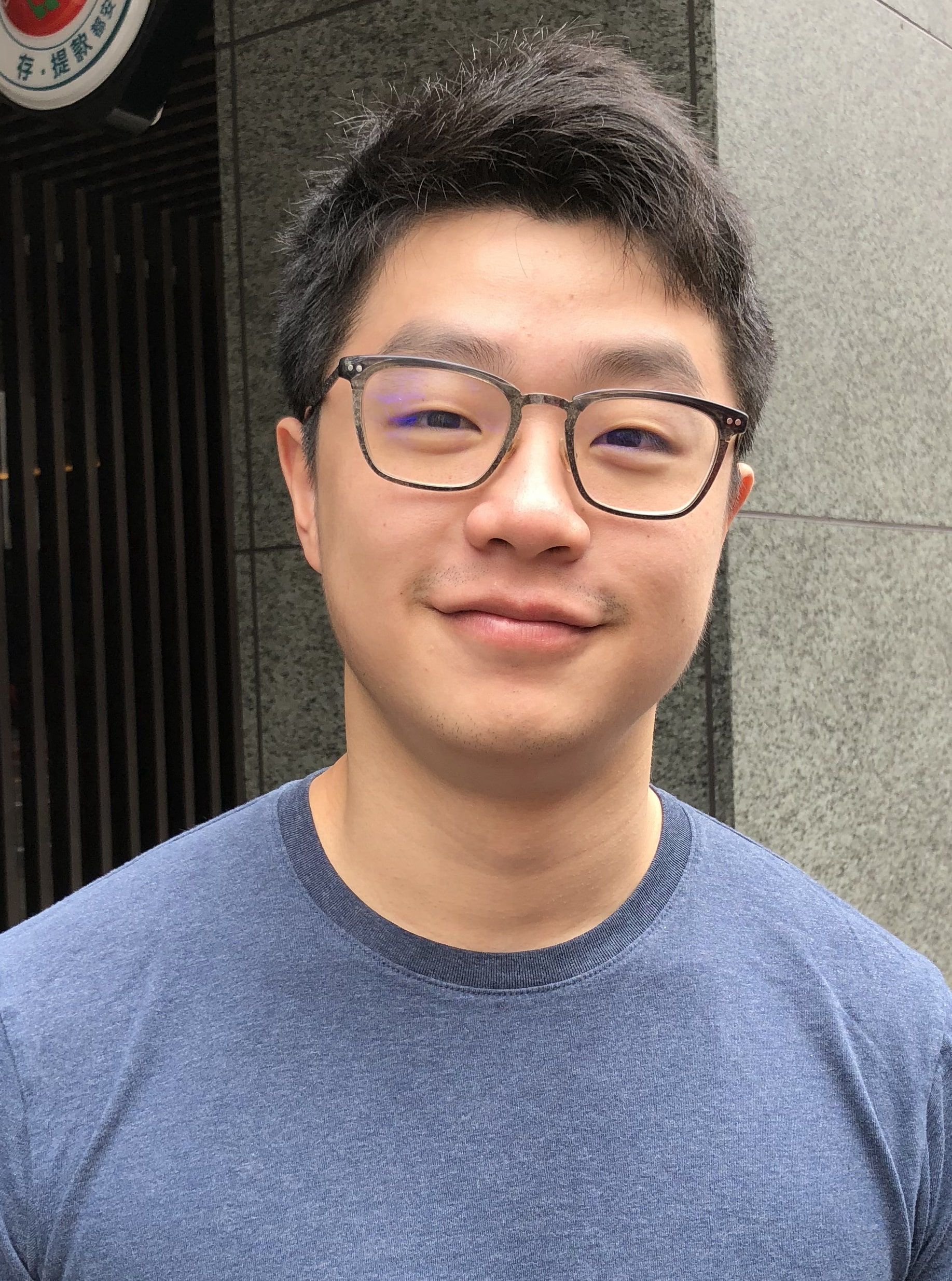}}]{Daxuan Ren}
is a PhD student in Computer Science at NTU advised by Prof. Jianmin Zheng (NTU) and Prof. Jianfei Cai (Monash University). Prior to his Ph.D, he was a software engineer in Autodesk working on large-scale 3D Reconstruction Engine (Autodesk ReCap Photo) including SFM, MVS, and Surface Reconstruction. His research interests include 3D Computer Vision and Graphics. He has multiple papers published at top-tier conferences including ICCV, and ECCV. 
  
\end{IEEEbiography}

\begin{IEEEbiography}[{\includegraphics[width=1in,height=1.25in,clip,keepaspectratio]{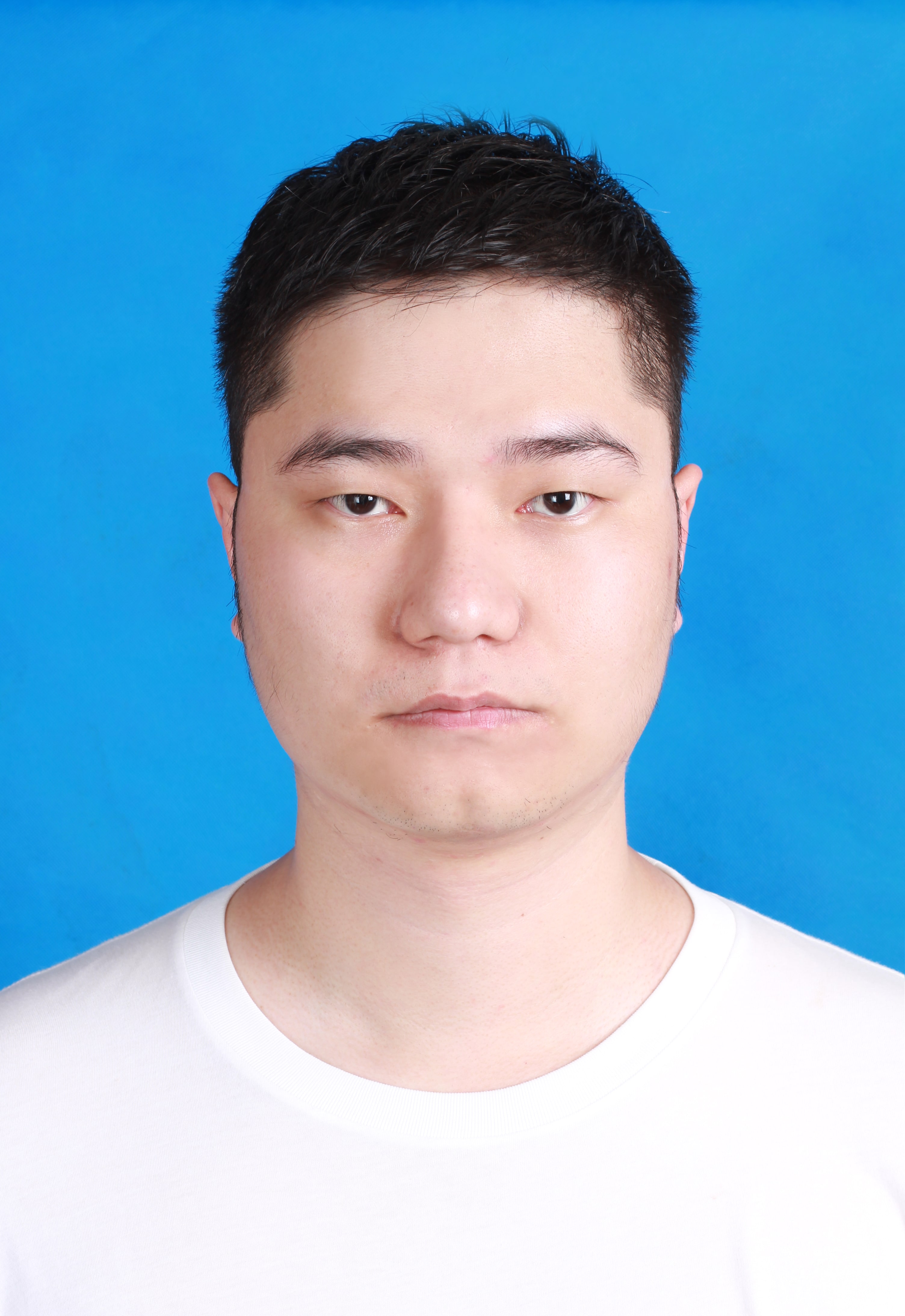}}]{Zhengyu Lin}
  is an Algorithm Engineer at SenseTime International Pte. Ltd. He attained his bachelor's degree at Imperial College London. His research interests include 3D computer vision and virtual data. He is currently working with human-centric RGB-D data.
\end{IEEEbiography}

\begin{IEEEbiography}[{\includegraphics[width=1in,height=1.25in,clip,keepaspectratio]{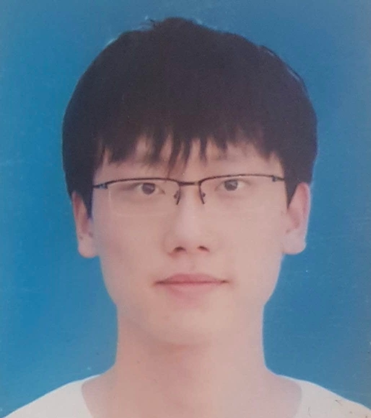}}]{Haiyu Zhao}
  is currently a senior researcher at SenseTime International Pte. Ltd. Previously, he was a research assistant at the Chinese University of Hong Kong, supervised by Prof. Xiaogang Wang and Prof. Hongsheng Li. Before that, Haiyu received his bachelor's degree at Beihang University. He has published over 10 papers (with more than 1,200 citations) at top-tier conferences in relevant fields, including CVPR, ICCV, ECCV, NeurIPS, and ICLR.
\end{IEEEbiography}

\begin{IEEEbiography}[{\includegraphics[width=1in,height=1.25in,clip,keepaspectratio]{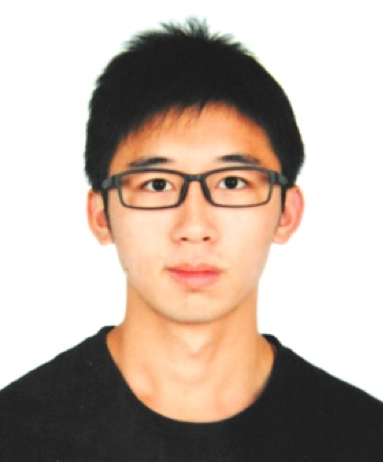}}]{Lei Yang}
  is currently a Research Director at SenseTime Group Inc., under the supervision of Prof. Xiaogang Wang. Prior to that, Lei received his Ph.D. degree from the Chinese University of Hong Kong in 2020, advised by Prof. Dahua Lin and Prof. Xiaoou Tang. Before that, Lei obtained his B.E. degree from Tsinghua University in 2015. He has published over 10 papers on top conferences in relevant fields, including CVPR, ICCV, ECCV, AAAI, SIGGRAPH and RSS.
\end{IEEEbiography}

\begin{IEEEbiography}[{\includegraphics[width=1in,height=1.25in,clip,keepaspectratio]{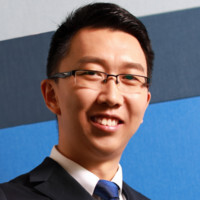}}]{Chen Change Loy}
 (Senior Member, IEEE) received the PhD degree in computer science from the Queen Mary University of London, in 2010. He is a professor with the College of Computing and Data Science , Nanyang Technological University. Prior to joining NTU, he served as a research assistant professor with the Department of Information Engineering, The Chinese University of Hong Kong, from 2013 to 2018. His research interests include computer vision and deep learning with a focus on image/video restoration and enhancement, generative tasks, and representation learning. He serves as an Associate Editor of the Computer Vision and Image Understanding (CVIU), International Journal of Computer Vision (IJCV) and IEEE Transactions on Pattern Analysis and Machine Intelligence (TPAMI). He also serves/served as the Area Chair of top conferences such as ICCV, CVPR, ECCV, ICLR and NeurIPS.
\end{IEEEbiography}

\begin{IEEEbiography}[{\includegraphics[width=1in,height=1.25in,clip,keepaspectratio]{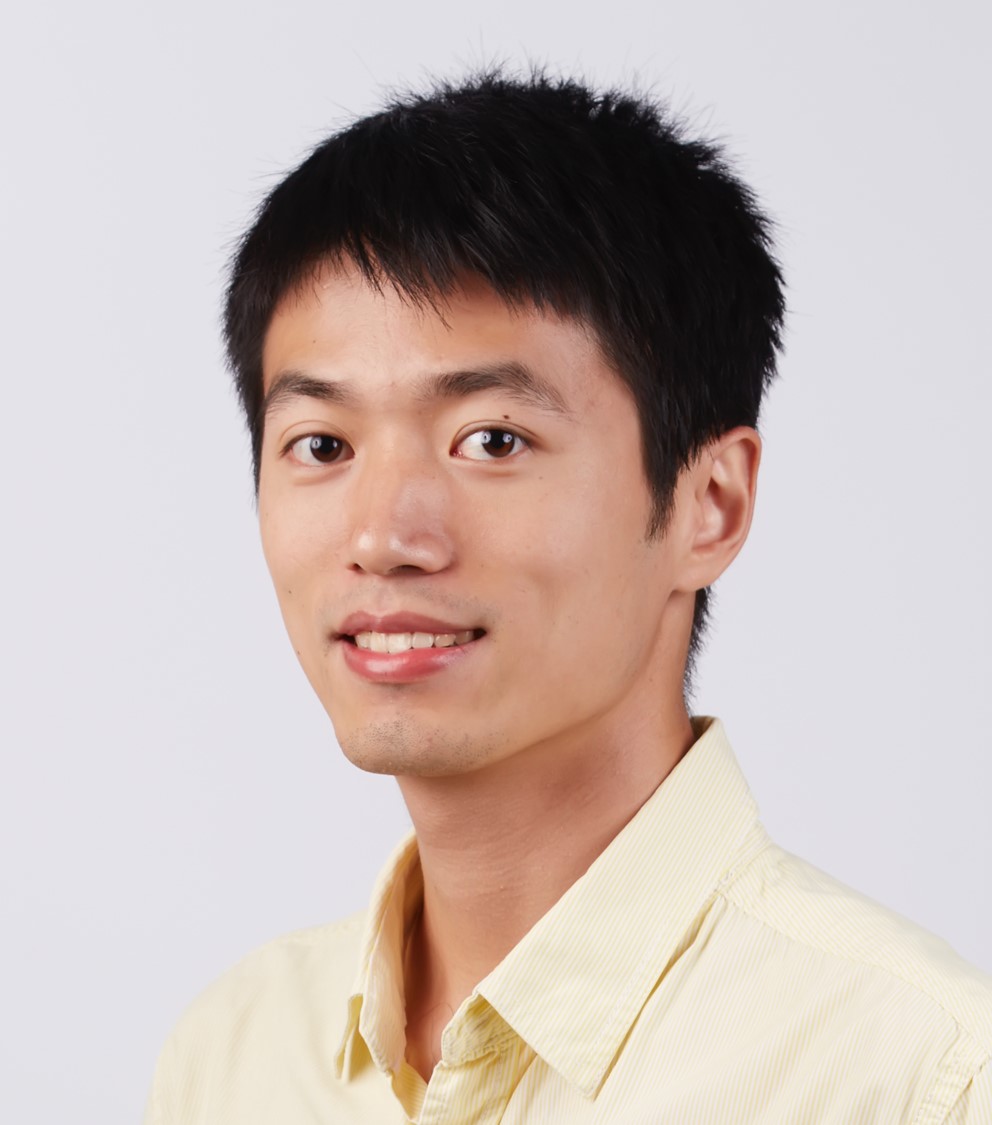}}]{Ziwei Liu}
  is currently an Assistant Professor at Nanyang Technological University, Singapore. Previously, he was a senior research fellow at the Chinese University of Hong Kong and a postdoctoral researcher at University of California, Berkeley. Ziwei received his PhD from the Chinese University of Hong Kong. His research revolves around computer vision, machine learning and computer graphics. He has published extensively on top-tier conferences and journals in relevant fields, including CVPR, ICCV, ECCV, NeurIPS, ICLR, ICML, TPAMI, TOG and Nature - Machine Intelligence. He is the recipient of Microsoft Young Fellowship, Hong Kong PhD Fellowship, ICCV Young Researcher Award and HKSTP Best Paper Award. He also serves as an Area Chair of ICCV, NeurIPS, and ICLR.
\end{IEEEbiography}

\vfill

\end{document}